\documentclass[letterpaper, 10 pt, journal, twoside]{IEEEtran}

\usepackage[only-used=true, single=false]{acro}
\usepackage[ruled,vlined,linesnumbered]{algorithm2e}
\usepackage{amsfonts, amsmath, amssymb}
\usepackage{array}
\usepackage{bm}
\usepackage{cite}
\usepackage{color}
\usepackage{float}
\usepackage[utf8]{inputenc} 
\usepackage[T1]{fontenc}    
\usepackage{graphicx}
\usepackage{subfig}
\usepackage{url}
\usepackage{mathtools}

\usepackage{tikz-cd}
\usetikzlibrary{fit, patterns}
\usetikzlibrary{shapes.geometric,fit}
\tikzset{
   dashellipse/.style={ellipse,draw,dashed,inner sep=0pt,blue,fit={#1}}
}
\pgfdeclarelayer{background}
\pgfsetlayers{background,main}

\usetikzlibrary{shapes.arrows}
\tikzset{
myarrow/.style={
  draw,thick,
  single arrow,
  text width=1cm,
  fill=white
  },
}

\let\oldnl\nl
\newcommand{\nonl}{\renewcommand{\nl}{\let\nl\oldnl}}
\newcommand{\forceindent}{\parindent=0.5em\indent\parindent=0pt\relax}
\newcommand{\algspacing}{\vspace{0.15cm}}

\usepackage{multirow}
\usepackage{booktabs}

\usepackage{dblfloatfix}    

\usepackage{adjustbox}

\title{\fontsize{16.5}{19.8} \bf Path Planning for Manipulation using Experience-driven Random Trees}

\author{{\`E}ric Pairet$^{1,2}$, Constantinos Chamzas$^{2}$, Yvan Petillot$^{1}$, Lydia E. Kavraki$^{2}$
\thanks{Manuscript received: October 8, 2020; Revised January 14, 2019; Accepted February 14, 2021. This paper was recommended for publication by Editor Hong Liu upon evaluation of the Associate Editor and Reviewers' comments.}%
\thanks{This research has been partially supported by the Scottish Informatics and Computer Science Alliance (SICSA), ORCA Hub EPSRC (EP/R026173/1) and consortium partners. Work by LEK and CC is supported in part by NSF 1718478 and NSF 2008720, Rice University Funds, and NSF 1842494 (CC).}%
\thanks{$^{1}$Edinburgh Centre for Robotics, University of Edinburgh and Heriot-Watt University (UK). {\tt\scriptsize eric.pairet@ed.ac.uk}, {\tt\scriptsize y.r.petillot@hw.ac.uk}}%
\thanks{$^{2}$Kavraki Lab, Department of Computer Science at Rice University, Houston TX (USA). {\tt\scriptsize chamzas@rice.edu}, {\tt\scriptsize kavraki@rice.edu}}%
\thanks{Digital Object Identifier (DOI): see top of this page.}
}

\definecolor{amber}{rgb}{1.0, 0.75, 0.0}

\newcommand*{\sref}[1]{Section~\ref{#1}}

\newcommand*{\aref}[1]{Algorithm~\ref{#1}}
\newcommand*{\eref}[1]{Equation~\ref{#1}}
\newcommand*{\fref}[1]{Figure~\ref{#1}}

\newcommand*{\lref}[1]{line~\ref{#1}}
\newcommand*{\lalref}[2]{line~\ref{#1}~and~\ref{#2}}
\newcommand*{\ltlref}[2]{line~\ref{#1}~to~\ref{#2}}

\SetKwRepeat{KwRepeatUntil}{repeat}{until}
\SetKw{KwDo}{do}
\SetKw{KwNot}{not}
\SetKw{KwReturn}{return}
\SetKw{KwAnd}{and}
\SetKw{KwOr}{or}
\newcommand{\method}[2]{\textup{\small#1}{\textup{(#2)}}}

\let\originalleft\left
\let\originalright\right
\renewcommand{\left}{\mathopen{}\mathclose\bgroup\originalleft}
\renewcommand{\right}{\aftergroup\egroup\originalright}

\DeclareMathAlphabet{\mymathbb}{U}{BOONDOX-ds}{m}{n}

\newcommand{\real}{\mathbb{R}}

\newcommand{\cspace}{\mathcal{Q}}
\newcommand{\sspace}{\mathcal{S}}

\newcommand{\s}{\textbf{s}}
\newcommand{\q}{\textbf{q}}

\newcommand{\traj}{\xi}

\newcommand{\demo}{\xi_\mathcal{D}}
\newcommand{\tree}{\mathcal{T}}

\newcommand{\demosegment}{\psi_\mathcal{D}}

\newcommand{\democonfigurationbar}{\bar{\q}{{}_\mathcal{D}}}
\newcommand{\demophasebar}{\bar{\alpha}{{}_\mathcal{D}}}

\newcommand{\tsb}{\textbf{b}}
\newcommand{\tss}{\bm{\lambda}}
\newcommand{\tsp}{\bm{\rho}}
\newcommand{\tse}{\bm{\epsilon}}
\newcommand{\tsg}{p}

\newcommand{\psl}{\omega_\alphamin}
\newcommand{\psu}{\omega_\alphamax}

\newcommand{\dist}{\text{dist}}

\newcommand{\free}{\text{free}}
\newcommand{\obst}{\text{obst}}

\newcommand{\alphamin}{\text{min}}
\newcommand{\alphamax}{\text{max}}

\newcommand{\ini}{\text{ini}}
\newcommand{\fin}{\text{end}}

\newcommand{\start}{\text{start}}
\newcommand{\goal}{\text{goal}}
\newcommand{\init}{\text{init}}

\newcommand{\targ}{\text{targ}}
\newcommand{\near}{\text{near}}

\usepackage[normalem]{ulem}
\definecolor{orange}{RGB}{255,128,0}
\definecolor{blue}{RGB}{0,128,255}

\newcommand{\modified}[1]{#1}

\usepackage{soul,xcolor}

\DeclareAcronym{1D}{
  short = 1D,
  long  = one-dimensional
}
\DeclareAcronym{2D}{
  short = 2D,
  long  = two-dimensional
}
\DeclareAcronym{3D}{
  short = 3D,
  long  = three-dimensional
}
\DeclareAcronym{AI}{
  short = AI,
  long  = artificial intelligence
}
\DeclareAcronym{CAN}{
  short = CAN,
  long  = controller area network
}
\DeclareAcronym{CF}{
  short = CF,
  long  = coupling force
}
\DeclareAcronym{CMP}{
  short = CMP,
  long  = compliant movement primitive
}
\DeclareAcronym{DMP}{
  short = DMP,
  long  = dynamic movement primitive
}
\DeclareAcronym{DoF}{
  short = DoF,
  long  = degree of freedom,
  long-plural-form = degrees of freedom
}
\DeclareAcronym{DS}{
  short = DS,
  long  = dynamical system
}
\DeclareAcronym{EM}{
  short = EM,
  long  = expectation-maximisation
}
\DeclareAcronym{ERT}{
  short = ERT,
  long  = experience-driven random trees
}
\DeclareAcronym{FK}{
  short = FK,
  long  = forwad kinematics
}
\DeclareAcronym{GMM}{
  short = GMM,
  long  = Gaussian mixture model
}
\DeclareAcronym{GMR}{
  short = GMR,
  long  = Gaussian mixture regression
}
\DeclareAcronym{GPR}{
  short = GPR,
  long  = Gaussian process regression
}
\DeclareAcronym{HMM}{
  short = HMM,
  long  = hidden Markov model
}
\DeclareAcronym{HRI}{
  short = HRI,
  long  = human-robot interaction
}
\DeclareAcronym{HSMM}{
  short = HSMM,
  long  = hidden semi-Markov model
}
\DeclareAcronym{KL}{
  short = KL,
  long  = Kullback-Leibler
}
\DeclareAcronym{IGMM}{
  short = IGMM,
  long  = infinite Gaussian mixture model
}
\DeclareAcronym{IIT}{
  short = IIT,
  long  = Italian Institute of Technology
}
\DeclareAcronym{IK}{
  short = IK,
  long  = inverse kinematics
}
\DeclareAcronym{ILC}{
  short = ILC,
  long  = iterative learning control
}
\DeclareAcronym{IR}{
  short = IR,
  long  = infra-red
}
\DeclareAcronym{LbD}{
  short = LbD,
  long  = learning by demonstration
}
\DeclareAcronym{LED}{
  short = LED,
  long  = light-emitting diode
}
\DeclareAcronym{LMS}{
  short = LMS,
  long  = least mean squares
}
\DeclareAcronym{LS}{
  short = LS,
  long  = linear square
}
\DeclareAcronym{LWPR}{
  short = LWPR,
  long  = locally weighted projection regression
}
\DeclareAcronym{LWR}{
  short = LWR,
  long  = locally weighted regression
}
\DeclareAcronym{MTR}{
  short = MTR,
  long  = multiple target regression
}
\DeclareAcronym{NMSE}{
  short = NMSE,
  long  = normalised mean squared error
}
\DeclareAcronym{NN}{
  short = NN,
  long  = neural network
}
\DeclareAcronym{OMPL}{
  short = OMPL,
  long  = the Open Motion Planning Library
}
\DeclareAcronym{OROCOS}{
  short = OROCOS,
  long  = open robot control software
}
\DeclareAcronym{PD}{
  short = PD,
  long  = proportional-derivative
}
\DeclareAcronym{RBF}{
  short = RBF,
  long  = radial basis function
}
\DeclareAcronym{RC}{
  short = RC,
  long  = regressor chain
}
\DeclareAcronym{ReLu}{
  short = ReLu,
  long  = rectified linear unit
}
\DeclareAcronym{RFWR}{
  short = RFWR,
  long  = receptive field weighted regression
}
\DeclareAcronym{RL}{
  short = RL,
  long  = reinforcement learning
}
\DeclareAcronym{ROS}{
  short = ROS,
  long  = robot operating system
}
\DeclareAcronym{RRT}{
  short = RRT,
  long  = rapidly-exploring random tree
}
\DeclareAcronym{RRTConnect}{
  short = RRTConnect,
  long  = bi-directional rapidly-exploring random tree
}
\DeclareAcronym{STR}{
  short = STR,
  long  = single target regression
}
\DeclareAcronym{WP}{
  short = WP,
  long  = work package
}
\DeclareAcronym{YARP}{
  short = YARP,
  long  = yet another robotic platform
}

\begin{document}
    \markboth{IEEE Robotics and Automation Letters. Preprint Version. Accepted February, 2021}{Pairet \MakeLowercase{\textit{et al.}}: Path Planning for Manipulation using Experience-driven Random Trees}

    \maketitle

    \begin{abstract}

    Robotic systems may frequently come across similar manipulation planning problems that result in similar motion plans. Instead of planning each problem from scratch, it is preferable to leverage previously computed motion plans, i.e., experiences, to ease the planning. Different approaches have been proposed to exploit prior information on novel task instances. These methods, however, rely on a vast repertoire of experiences and fail when none relates closely to the current problem.
    Thus, an open challenge is the ability to generalise prior experiences to task instances that do not necessarily resemble the prior.
    This work tackles the above challenge with the proposition that experiences are ``decomposable'' and ``malleable'', i.e., parts of an experience are suitable to relevantly
    explore the connectivity of the robot-task space even in non-experienced regions. Two new planners result from this insight: \acf{ERT} and its bi-directional version ERTConnect. These planners adopt a tree sampling-based strategy that incrementally extracts and modulates parts of a single path experience to compose a valid motion plan.
    We demonstrate our method on task instances that significantly differ from the prior experiences, and \mbox{compare with 
    related} state-of-the-art experience-based planners.
    While their
    repairing strategies fail to
    generalise priors of tens of experiences, our planner, with a single experience, significantly outperforms them in both success rate and planning time. 
    Our planners are implemented and freely available in \acl{OMPL}.
\end{abstract}


\begin{IEEEkeywords}
    Manipulation Planning; Motion and Path Planning; Learning from Experience; Autonomous Agents
\end{IEEEkeywords}

    \section{INTRODUCTION} \label{sec:introduction}

\begin{figure}[t]
        \centering
        \includegraphics[width=0.98\columnwidth]{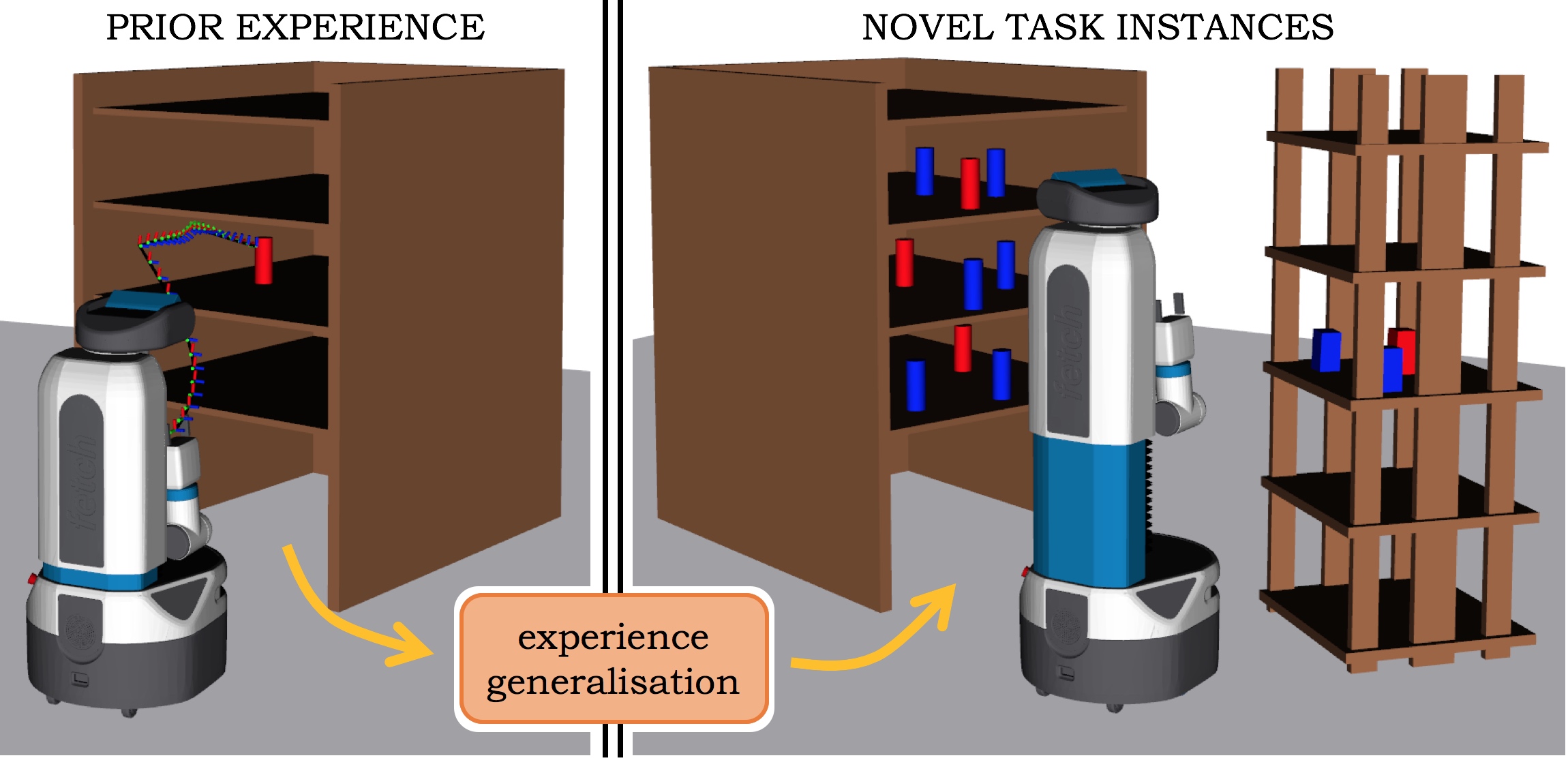}
        \caption{
        Our planner can leverage a single path (experience) computed in a particular task instance, e.g., ``fetch the red object'' (left), to efficiently solve novel task instances (right) that remarkably differ from the experience, e.g., in obstacles (blue objects), shelf structural geometry and target locations.}
        \label{fig:introduction}
    \end{figure}

    A long-envisioned requisite for fully-autonomous robotic manipulation is to endow robots with the ability to learn from and improve through experiences. For example, consider a robot on a shelf stacking task (see \fref{fig:introduction}). Such a robot may frequently come across similar task instances that result in similar motion plans. Despite the resemblance among problems, the most common approach is to plan from scratch; neither prior information nor recurrent computations are leveraged to aid in solving related queries. This strategy can lead to unnecessary long planning times. Instead, the commonalities between instantiations should be considered as prior knowledge at the planning stage. However, this is not a trivial problem. The planner must reason over the relevant features that allow for the generalisation of prior knowledge even to notably different task instances.

    \textbf{Related work.} Leveraging prior experiences for building motion plans efficiently has drawn special attention to the learning and planning communities. Learning-based approaches infer the underlying task policy from a given set of demonstrations, which is then used to retrieve task-related motion plans (e.g.,~\cite{meier2018online,pairet2019blearning,stark2019experience,pairet2019alearning}). Relevant features are extracted from the demonstrations such that the learnt policy can generalise to novel task instances. Although these methods are capable of computing plans quickly
    by learning from experience, they typically generalise poorly to task instances that significantly differ from those observed a priori~\cite{ravichandar2020recent}.

    On coping with varying task instances while leveraging experiences, sampling-based planning offers a promising venue to generalise the a priori knowledge. Such a strategy is known as experience-based planning. There are mainly two orthogonal approaches: (1)~biasing the sampling into task-relevant areas, and (2)~exploiting previously computed motions. This work is, in spirit, closer to the latter: leveraging prior motions. Related work is discussed for both alternatives.
    
    (1)~Biasing the sampling involves guiding the exploration towards task-relevant regions of the configuration space. A common approach is to take advantage of geometric features of the workspace to guide the sampling in the configuration space (e.g.,~\cite{zucker2008adaptive,ichter2018learning,lehner2018repetition,chamzas2019using,molina2020learn}). Strategies that bias the sampling can significantly speed up queries, but they rely on identifying familiar workspace features to infer relevant samples in the configuration space. Therefore, their applicability is mainly limited to task instances that resemble those observed a priori, leading to a lack of generalisation to new environments.
    {\let\thefootnote\relax\footnote{A visual aid about the \acl{ERT} planners can be found in: \texttt{\url{https://youtu.be/kD3A3Xs_psI}}.}}

    (2)~Using previously computed motions consists of storing experienced motions in a library (e.g.,~\cite{berenson2012robot,jetchev2013fast})~or~jointly as a graph (e.g.,~\cite{phillips2012graphs,coleman2015experience}). These methods recall exact prior experiences to solve the current planning query. In Lightning~\cite{berenson2012robot}, the most relevant experience is retrieved based on the start-goal proximity of a experience and the current query. The nearest path is chosen to be repaired. The repair step employs the \ac{RRTConnect} to re-connect the end-points of segments originated by variant constraints, e.g., obstacles. 
    Differently, Experience Graphs~\cite{phillips2012graphs} build a roadmap of experiences to then search it 
    using some heuristics. In a similar vein, Thunder~\cite{coleman2015experience} creates a sparse roadmap from all experiences, which is repeatedly queried via $\text{A\!}^*$ until a valid path is found. If the graph does not contain a valid path, candidate paths, if any, are considered for repairing. The repairing invokes RRTConnect to reconnect the disconnected states along the candidate paths. All these path-centric approaches exploit prior motion plans in the exact configuration they were experienced, i.e., ``rigidly''. This leads to poor performance when planning in non-experienced regions of the robot-task space. Therefore, for these methods to work, the \mbox{library must contain a prior path that already} resembles a valid motion plan for the current planning problem. Consequently, current experienced-based planners are dependant on a vast and extremely relevant set of prior experiences to counteract their lack of generalisation capabilities.

    \textbf{Contribution.} In this work, we change the paradigm in which prior path experiences stored in libraries of motions are being used. Instead of exploiting prior motions ``rigidly'' to preserve the invariant constraints, we use them in a \mbox{``decomposable''} and ``malleable'' way to infer the next move given a particular state of the robot in a task.
    With this proposition, we present \acf{ERT} and its bi-directional version ERTConnect, two experience-based planners capable of generalising a single prior motion plan across significantly varied task instances. These planners leverage a path experience by parts to iteratively build a tree of micro-experiences, i.e., segments that resemble those in the prior experience. Suitable micro-experiences result from
    semi-randomly morphing different parts of the experience.
    Such a strategy proves to be useful to efficiently explore the connectivity of the robot-task space even in non-experienced regions. 
    Additionally, we discuss how to select the best candidate for our planner given a library of path experiences. Such experience selection strategy enables the use of our planners in frameworks that incrementally build libraries of experiences by adding newly computed motion plans (e.g.,~\cite{berenson2012robot,coleman2015experience}), as well as in systems that gather experiences from human demonstrations (e.g.,~\cite{wang2019motion,delpreto2020helping,ardon2020self}). 
    
    The key insight of our approach is that prior experiences are of a better use when leveraged in a ``malleable'' fashion, oppositely to the common ``rigid'' usage of experiences. Thus, contrary to prior work, the applicability of our planner goes beyond task instances that closely resemble those observed a priori. Empirical analysis demonstrates our planner's ability to leverage prior experiences efficiently and to generalise them to distinguishably dissimilar task instances. In these challenging conditions, while related state-of-the-art experience-based planners fail to exploit vast repertoires of prior path experiences, our planner, with a single path experience, significantly outperforms them in both success rate and planning time.



    

    

    
    
    

    

\section{PROBLEM DEFINITION and \\ APPROACH OVERVIEW} \label{sec:preliminaries}
    
        In this manuscript, we are interested in families of motion planning problems that involve similar task instances and thus, seek similar motion plans. The commonalities between instantiations are of interest because they open the possibility for a robot to leverage prior information about the task. Enabling the robot to exploit such similarities would allow it to efficiently solve tasks related to those seen a priori. 
        

            
        Consider a robot with configuration space ${\cspace \in \real^n}$ conducting a particular task, e.g., shelf stacking (see \fref{fig:introduction}). Let ${\cspace_\obst \subset \cspace}$ be the region of the configuration space occupied by obstacles, and ${\cspace_\free = \cspace \setminus \cspace_\obst}$ be the collision-free region. 
        Let ${\q \in \cspace}$ denote a particular robot configuration, and ${\alpha \in [0,1]}$ be a phase variable that indicates the progress on the execution of a collision-free motion plan. Then, 
        the state of the robot in a motion plan is defined in the configuration-phase space $\sspace = \cspace \times \real_{[0,1]}$ as ${\s = \langle \q, \alpha \rangle} \in \real^{n+1}$. 
        The valid regions in the configuration-phase are defined as:
        \begin{align}
            \sspace_\free = \{ \langle \q, \alpha \rangle \in \sspace \; | \; \q \in \cspace_\free \}.
        \end{align}
        
        
        Let $\mathcal{A}$ be some prior information about the task. In this work, we consider prior knowledge defined by a library of path experiences ${\mathcal{A} = \{{\demo}_1, {\demo}_2, \dots, {\demo}_j\}}$, where each ${\demo}$ is a path (prior experience) solving a particular task instance. Paths as priors are of particular interest since they can be acquired over time from the robot's planning solutions on similar task instances, or from external sources, such as from a human kinaesthetically guiding a robot through a task. Note that the focus of this manuscript is experience-based planning, where a set of prior path experiences $\mathcal{A}$ relevant to the current problem is assumed to be provided.
        Therefore, given a library ${\mathcal{A}}$, and the start ${\langle \q_\start, 0 \rangle \in \sspace_\free}$ and goal ${\langle \q_\goal, 1 \rangle \in \sspace_\free}$ states, the motion planning problem considered in this work seeks a planning process ${\mathcal{J}: \mathcal{A} \rightarrow \xi}$ capable to leverage $\mathcal{A}$ to efficiently find a collision-free continuous path ${\traj: \alpha \in [0,1] \rightarrow \sspace_\free}$ that connects
        ${\traj(0) = \q_{\start} \in \sspace_\free}$ to 
        ${\traj(1) = \q_{\goal} \in \sspace_\free}$. 

        Our approach to take advantage of a library of experiences ${\mathcal{J}: \mathcal{A} \rightarrow \xi}$ is twofold. First, as discussed in \sref{sec:preliminaries_selection}, we select a path experience ${\demo \in \mathcal{A}}$ suitable for the current planning problem.
        Then, we exploit the selected prior ${\mathcal{L}: \demo \rightarrow \xi}$ via our contribution: the experience-driven random trees planners ERT and ERTConnect presented in \sref{sec:planner}. We empirically demonstrate that, when using our planner, a unique prior path suffices to solve other instances of the same task.

	\section{EXPERIENCE-DRIVEN RANDOM TREES} \label{sec:planner}
    
    
    The ERT and ERTConnect planners are inspired by tree sampling-based methods~\cite{hsu1997path,kuffner2000rrt}. Our planners, however, iteratively leverage a single task-relevant prior path experience by parts (segments, a.k.a., micro-experiences) to ease the capture of connectivity of the space. Such micro-experiences are semi-randomly morphed to generate task-relevant motions, i.e., segments that resemble those in the prior (e.g., dotted lines in \fref{fig:morphing}). The obtained motions are sequentially concatenated to compose a task-relevant tree (see green tree in \fref{fig:illustration_connect_explore}). This exploratory strategy aims at finding a trace along the tree edges, i.e., a sequence of local modifications on the prior, that constitutes a 
    continuous path $\traj$ which satisfies ${\traj: \alpha \in [0,1] \rightarrow \sspace_\free}$, ${\traj(0) = \q_\start}$ and ${\traj(1) = \q_\goal}$.
    
    Noteworthy, our planners are designed to be agnostic to distance metrics, as capturing proximity between two robot configurations in a task is not trivial. Moreover, such metric would potentially need to be designed for each task. Therefore, instead of iteratively growing a tree from the nearest configuration to a random sample (RRT-like~\cite{kuffner2000rrt}), our \acl{ERT} iteratively branch-off (expand, EST-like~\cite{hsu1997path}) by concatenating the inferred motions. Likewise, to generate resembling motions, we deform the micro-experiences such that no similarity metric is needed.

    The core routine through which the planner exploits the prior experience to generate task-relevant micro-experiences is detailed in \sref{sec:planner_segment}, and its usage in a uni- and bi-directional sampling-based planning strategy is presented in \sref{sec:planner_uni} and \sref{sec:planner_bi}, respectively.

    \subsection{Inferring Task-relevant Motions from a Single Experience \label{sec:planner_segment}}%
    
        
        For planning efficiently, we are particularly interested in generating motions that are task-relevant in $\sspace$, i.e., coherent according to the robot state in a task. To that purpose, our planners leverage an experience by parts in a ``malleable'' fashion, as opposed to the common ``rigid'' usage, to infer motions that are likely to be relevant to different task instantiations.
        
\begin{figure}[t]
            \centering
            \includegraphics[width=\columnwidth]{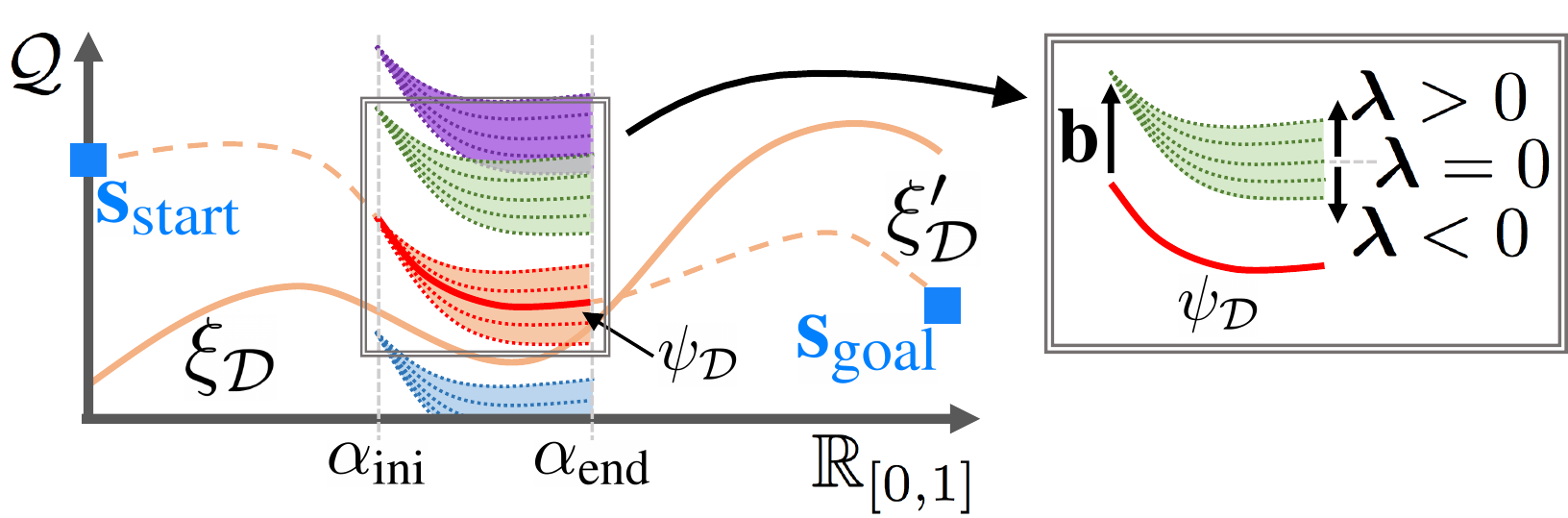}
            \caption{Illustrative example of \eref{eq:segment_transform}: generation of resembling motions (dotted lines) by morphing the micro-experience~$\demosegment$ with semi-random $\tsb$ (shift) and $\tss$ (shear) pairs.
            }
            \label{fig:morphing}
        \end{figure}

        Initially, our planners pre-process the given experience $\demo$ before exploiting it iteratively. Specifically, $\demo$ is mapped onto the current planning problem to obtain $\demo^\prime$, a path whose initial and final configurations match the start and goal of the current planning problem (see \fref{fig:morphing}). The computation of such mapping ${\demo \rightarrow \demo^\prime}$ is detailed within the description of the planners.
        Then, at each iteration, our planners leverage a part (micro-experience) of the mapped experience~$\demo^\prime$ to infer suitable motions for the task. Generally, let ${\demosegment: \alpha \in [\alpha_{\ini},\alpha_{\fin}]}$ be a micro-experience from the prior spanning from $\alpha_{\ini}$ to $\alpha_{\fin}$ such that ${\demosegment(\alpha) = \demo^\prime(\alpha) \; \forall \; \alpha \in [\alpha_{\ini},\alpha_{\fin}]}$ (e.g., red segment in \fref{fig:morphing}). We denote the extraction of a micro-experience from a prior as ${\demosegment = \demo^\prime(\alpha_{\ini},\alpha_{\fin})}$, and say that such segment has a phase span ${|{\demosegment}_{\alpha}| \in (0,1]}$.
        
        
        Extracted micro-experiences are exploited to create task-relevant motions. 
        Formally, let ${\nu: \demosegment \rightarrow \psi \in \real^{(n+1) \times (n+1)}}$ be a function that morphs a sequence of states onto another region of $\sspace$. We formulate the support of this operation to be that of an affine transformation of the form ${\psi = A\demosegment + B}$, where ${{\demosegment}_{(n+1)\times k} = \langle {\democonfigurationbar}_{n \times k}, {\demophasebar}_{1 \times k} \rangle}$ is a prior micro-experience with $k$ states, and ${\psi_{(n+1)\times k}= \langle \bar{\q}_{n \times k}, \bar{\alpha}_{1 \times k} \rangle}$ is the generated task-relevant segment. Specifically, we design $A$ to be a shear transform
        for its shape-preserving properties, and $B$ to be a translation of the segment into a region of interest. Formally, then, this affine transformation modulates $\demosegment$ as:
        \begin{align}
            \!\!\!\!\!\!\!\!\!\!\! 
            \begin{bmatrix}
                \bar{\q} \\
                \bar{\alpha}
            \end{bmatrix} &=
            \begin{bmatrix}
                \mathbb{I}_{n \times n} & \tss_{n \times 1} \\
                \mymathbb{0}_{1 \times n} & |{\demosegment}_\alpha|
            \end{bmatrix}
            \begin{bmatrix}
                \democonfigurationbar \\
                \tsp
            \end{bmatrix} + 
            \begin{bmatrix}
                \tsb_{n \times 1} & \!\!\!\! \dots \!\!\!\! & \tsb_{n \times 1} \\
                \alpha_\ini & \!\!\!\! \dots \!\!\!\! & \alpha_\ini
            \end{bmatrix}_{(n + 1) \times k}
            \!\!\!\!\!\!\!\!\!\!\!\!\!\!\!\!\!\!\!, 
            \quad 
          \label{eq:segment_transform}
        \end{align}
        where $\tss_{n \times 1}$ is the shearing coefficient, $\tsb_{n \times 1}$ is a shifting vector, and ${\tsp = [0,...,1]_{1 \times k}}$ is a local reparametrisation of $\demophasebar$. Note that the phase of the generated segment $\psi$ remains ${\bar{\alpha} = \demophasebar}$. 
        Informally, \eref{eq:segment_transform} translates and smoothly deforms the micro-experience by adding up the increasing amount of noise ${\tss \tsp + \tsb}$, such that ${\psi(\alpha_\ini) - \demosegment(\alpha_\ini) = \tsb}$ and ${\psi(\alpha_\fin) - \demosegment(\alpha_\fin) = \tss + \tsb}$. Therefore, specifying $\tsb$ and $\tss$ enables the generation of new micro-experiences and their mapping onto any region of interest in $\sspace$. \fref{fig:morphing} exemplifies the affine morphing in \eref{eq:segment_transform} with different parameters. We detail the implementation of \eref{eq:segment_transform} in \aref{alg:morph}.
        
        \begin{algorithm}[h!]
            \SetInd{0.5em}{0.5em}
            \DontPrintSemicolon
            
            \caption{\mbox{MORPH\_SEGMENT($\demosegment$, $\tss$, $\tsb$)}}
            \label{alg:morph}
            
            \nonl\textbf{Input:} \\
            \nonl\forceindent$\demosegment$: micro-experience of phase-span $|{\demosegment}_\alpha|$ from $\alpha_\ini$ \\
            \nonl\forceindent$\tss$: shearing coefficient \\
            \nonl\forceindent$\tsb$: shifting vector \\
            
            \nonl\textbf{Output:} \\
            \nonl\forceindent$\psi$: morphed motion
            
            \algspacing
            
            \For(\tcp*[f]{\eref{eq:segment_transform}}){$\rho \leftarrow 0$ \textbf{\textup{to}} $1$} {
                $\alpha = \rho|{\demosegment}_\alpha| + \alpha_\ini$ \\
                $\psi(\alpha) \leftarrow \demosegment(\alpha) + \rho\tss + \tsb$
            }
            \KwReturn $\psi$
        \end{algorithm}

    
        
\begin{figure*}[t!]
        \centering
        \subfloat[ERT: first iteration example]{\includegraphics[width=0.3\textwidth, trim=0.1cm 0.0cm 0.3cm 0.3cm, clip]{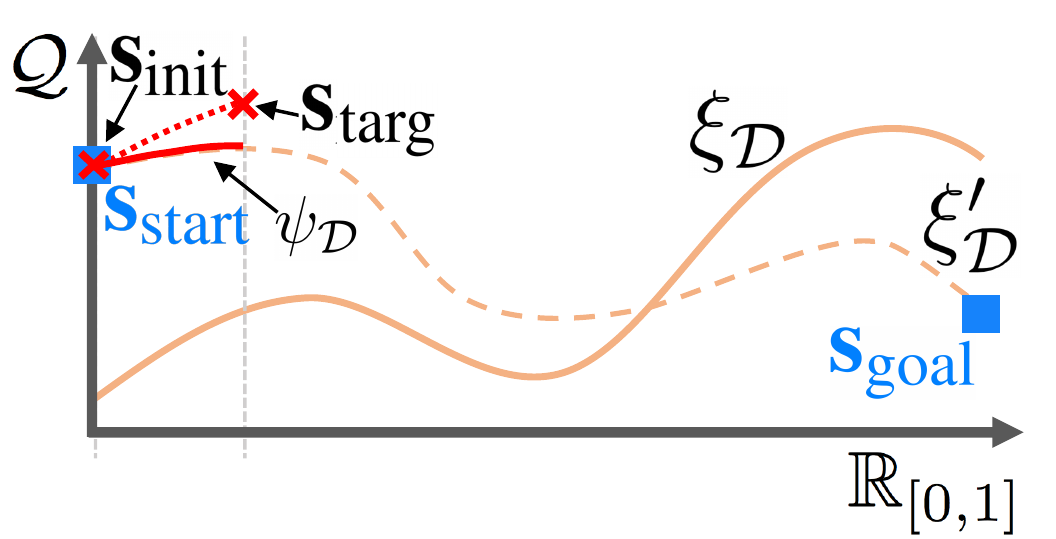}}
        \quad
        \subfloat[ERT: \textit{explore} example]{\includegraphics[width=0.3\textwidth, trim=0.1cm 0.0cm 0.3cm 0.3cm, clip]{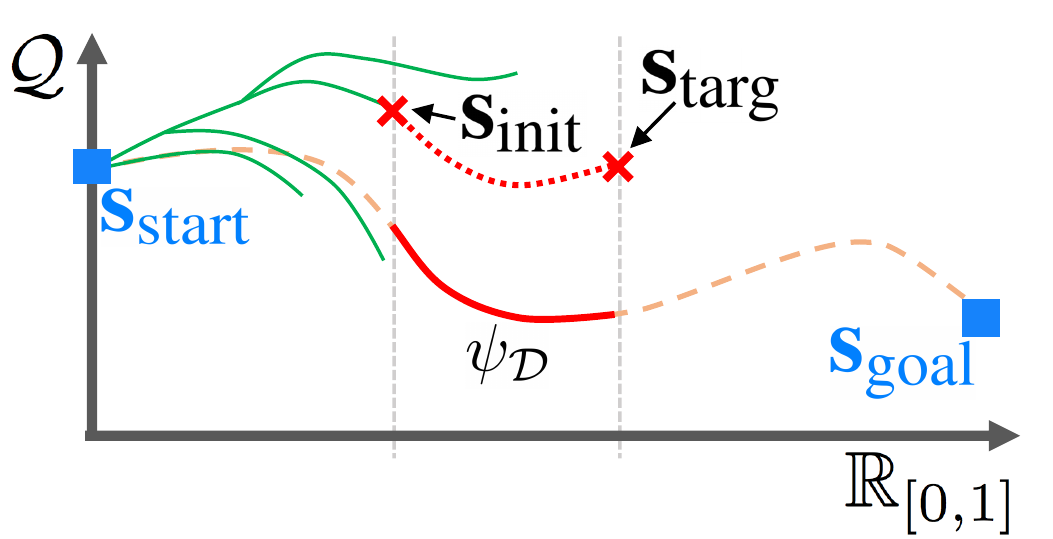}}
        \quad
        \subfloat[ERT: \textit{connect} example]{\includegraphics[width=0.3\textwidth, trim=0.1cm 0.0cm 0.3cm 0.3cm, clip]{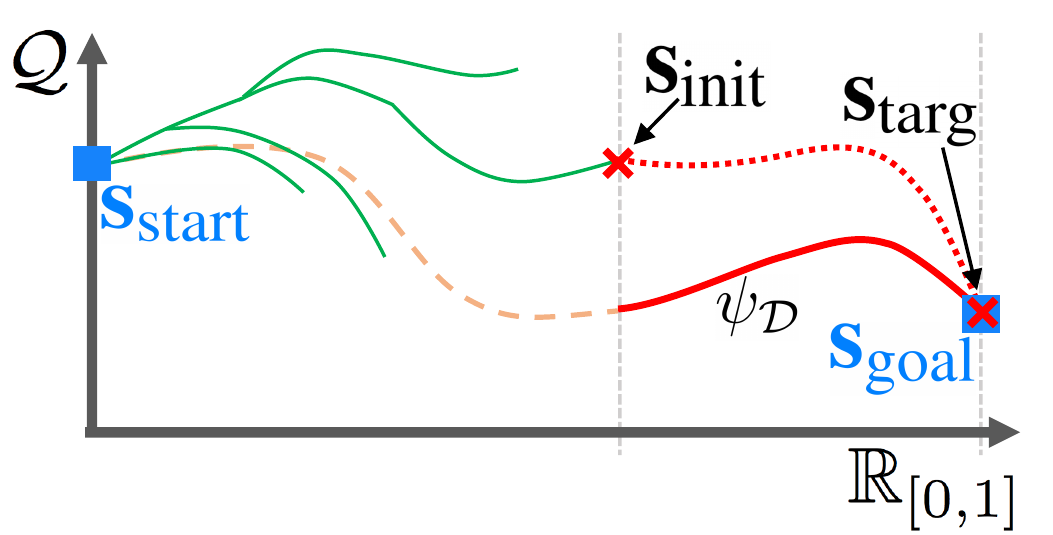}}
        \\
        \caption{Experience-driven random trees iteratively build a tree (green) of micro-experiences. At each iteration, an existing node in the tree is randomly selected to either \textit{explore} the most suitable continuation of the task (e.g., snapshots in (a) and (b)), or \textit{connect} to another known state (e.g., the goal state as in (c) (ERT), or a state in the other tree (ERTConnect)). In both cases, relevant motions (dotted red) are generated by morphing micro-experiences (red) of the prior path experience~$\demo^\prime$ (see~\fref{fig:morphing}).}
        
        \label{fig:illustration_connect_explore}
        \vspace{-0.35cm}
    \end{figure*}

        We exploit the ability to morph parts of the mapped prior path experience $\demo^\prime$ to infer motions that are suitable to either \textit{connect} two particular states $\s_\init$ and $\s_\targ$, or \textit{explore} the best way to continue the task from a given state $\s_\init$. These two processes, and their interaction with \aref{alg:morph}, are detailed in \aref{alg:segment} and illustrated in \fref{fig:illustration_connect_explore}.

        
        \begin{algorithm}[b!]
            \SetInd{0.5em}{0.5em}
            \DontPrintSemicolon
            
            
            \caption{\mbox{GENERATE\_SEGMENT($\s_\init$, $\s_\targ$, $\demo^\prime$)}}
            \label{alg:segment}
            
            \nonl\textbf{Input:} \\
            \nonl\forceindent$\s_\init$: required segment configuration-phase start \\
            \nonl\forceindent\mbox{$\s_\targ$: required (if any) segment configuration-phase end} \\
            \nonl\forceindent$\demo^\prime$: prior experience \\
            
            \nonl\textbf{Output:} \\
            \nonl\forceindent$\psi$: generated segment \\
            \nonl\forceindent$\s_\fin$: end configuration-phase of the segment $\psi$ \\
            
            \algspacing
            
            $\langle \q_\init, \alpha_\init \rangle = \s_\init$ \\

            \If(\tcp*[f]{connect}){\KwNot $\s_\targ = \varnothing$ \label{alg_line:gs_conditional}} {
                $\langle \q_\targ, \alpha_\targ \rangle = \s_\targ$ \label{alg_line:gs_connect_unpack} \\
                $\demosegment \leftarrow \demo^\prime(\alpha_\init$, $\alpha_\targ)$ \label{alg_line:gs_connect_extract} \\
                $\tsb \leftarrow \q_\init - \demosegment(\alpha_\init)$ \label{alg_line:gs_connect_tsb} \\
                $\tss \leftarrow \q_\targ - (\demosegment(\alpha_\targ) + \tsb)$ \label{alg_line:gs_connect_tss}
            }
            \Else(\tcp*[f]{explore}) {
                $\alpha_\targ \leftarrow$ \method{SAMPLE\_SEGMENT\_END}{$\alpha_\init$} \label{alg_line:gs_explore_sample_end} \\
                
                
                $\demosegment \leftarrow \demo^\prime(\alpha_\init, \; \alpha_\targ)$ \label{alg_line:gs_explore_extract} \\
                $\tsb \leftarrow \q_\init - \demosegment(\alpha_\init)$ \label{alg_line:gs_explore_tsb} \\
                $\tss \leftarrow \mathbb{U}(-\tse |{\demosegment}_\alpha|, \; \tse |{\demosegment}_\alpha|)$ \label{alg_line:gs_explore_tss}
            }
            $\psi \leftarrow$ \method{MORPH\_SEGMENT}{$\demosegment$, $\tss$, $\tsb$} \label{alg_line:gs_map} \\
            $\s_\fin \leftarrow \langle \psi(\alpha_\targ), \alpha_\targ \rangle$ \\
            \KwReturn $\langle\psi, \; \s_\fin\rangle$
        \end{algorithm}

        \textit{Connect} (\ltlref{alg_line:gs_connect_unpack}{alg_line:gs_connect_tss}): given two task-related configuration-phase states ${\s_\init = \langle \q_\init, \alpha_\init \rangle}$ and ${\s_\targ = \langle \q_\targ, \alpha_\targ \rangle}$, we hypothesise that a suitable connection 
        may result from mapping the micro-experience ${\demosegment: \alpha \in [\alpha_\init, \alpha_\targ]}$ between $\s_\init$ and $\s_\targ$. 
        Thus, after extracting the relevant micro-experience from 
        $\demo^\prime$ 
        (\lref{alg_line:gs_connect_extract}), the parameters $\tsb$ and $\tss$ of the mapping in \eref{eq:segment_transform} are calculated such that the resulting micro-experience~$\psi$ satisfies ${\psi(\alpha_\init) = \q_\init}$ and ${\psi(\alpha_\targ) = \q_\targ}$ (\lalref{alg_line:gs_connect_tsb}{alg_line:gs_connect_tss}).

        
        
        \textit{Explore} (\ltlref{alg_line:gs_explore_sample_end}{alg_line:gs_explore_tss}): given one task-related configuration-phase state ${\s_\init = \langle \q_\init, \alpha_\init \rangle}$, we hypothesise that a suitable continuation of the task is to apply a micro-experience similar to that ${\demosegment \in \demo^\prime}$ starting at $\alpha_\init$. For that, we first determine which span of $\demo^\prime$ to exploit by defining $\alpha_\targ$ (\lref{alg_line:gs_explore_sample_end}). An appropriate $\alpha_\targ$ depends on the direction in which $\demo^\prime$ is being exploited; we call it \texttt{forward} when exploiting the prior from $\demo^\prime(0)$ to $\demo^\prime(1)$, and \texttt{backward} otherwise. Correspondingly, \method{SAMPLE\_SEGMENT\_END}{$\alpha_\init$} defines $\alpha_\targ$ as:
        \begin{align} \raisetag{4.6ex}
            \!\!\!\!
            \resizebox{0.93\columnwidth}{!}{$%
            \alpha_\targ = 
            \begin{cases}
                \min\left(\alpha_\init + \mathbb{U}\left(\psl, \; \psu\right), \; 1\right), & \text{if} \; \texttt{forward} \\
                \max\left(0, \; \alpha_\init - \mathbb{U}\left(\psl, \; \psu\right)\right), & \text{if} \; \texttt{backward}
            \end{cases}
            $}
        \end{align}
        where $\mathbb{U}(\psl, \psu)$ draws a sample from a uniform distribution to determine the phase span of the extracted segment. The bounds $\psl$ and $\psu$ are discussed in \sref{sec:planner_parameters}. Next, the corresponding segment ${\demosegment: \alpha \in [\alpha_\init, \alpha_\targ]}$ is extracted from the mapped prior path experience $\demo^\prime$ (\lref{alg_line:gs_explore_extract}), and
        $\tsb$ is computed for the resulting micro-experience $\psi$ to start at $\s_\init$, i.e., to satisfy ${\psi(\alpha_\init) = \q_\init}$ (\lref{alg_line:gs_explore_tsb}). Finally, to generate a task-relevant motion from $\q_\init$, the shearing coefficient $\tss$ is sampled randomly within some bounds to morph the micro-experience $\demosegment$ into a similar motion (\lref{alg_line:gs_explore_tss}).
        In particular, $\tss$ is drawn from a uniform distribution $\mathbb{U}\left(-\tse |{\demosegment}_\alpha|, \; \tse |{\demosegment}_\alpha|\right)$ such that, at each iteration, the maximum allowed deformation is proportional to the segment's phase span. This implies that the accumulated deformation along any possible path $\traj$ found by our planners does not exceed, with respect to $\demo^\prime$, the user-defined malleability bound $\tse$ (see discussion in \sref{sec:planner_parameters}).

        Overall, the method \method{GENERATE\_SEGMENT}{$\cdot$} enables the presented experience-guided random tree planners to leverage a single path experience at different levels of granularity, and map task-relevant segments onto any region of interest in the configuration-phase space. In that way, our planner aims at composing a valid path from a suitable sequence of morphed micro-experiences. The remaining of this section discusses the usage of such routine in our uni-directional (ERT) and a bi-directional (ERTConnect) tree sampling-based techniques.

    \subsection{Uni-directional Experience-driven Random Trees (ERT) \label{sec:planner_uni}}
    
        \begin{algorithm}[b!]
            \SetInd{0.5em}{0.5em}
            \DontPrintSemicolon
            \caption{ERT($\s_\start$, $\s_\goal$, $\demo$)} \label{alg:evs}
            
            \nonl\textbf{Input:} \\
            \nonl\forceindent$\s_\start$ and $\s_\goal$: start and goal configuration-phase \\
            \nonl\forceindent$\demo$: prior experience \\
            
            \nonl\textbf{Output:} \\
            \nonl\forceindent$\xi$: collision-free path \\
            
            \algspacing
            \tcc{map $\demo$ onto current problem}
            $\langle\demo^\prime, \, \varnothing\rangle \leftarrow$ \method{GENERATE\_SEGMENT}{$\s_\start$, $\s_\goal$, $\demo$} \label{alg_line:evs_pro1} \\
            \If{\method{IS\_VALID}{$\demo^\prime$} \label{alg_line:evs_pro2}} {
                \KwReturn $\demo^\prime$
            }
            
            \algspacing
            \tcc{sampling-based $\demo^\prime$ exploitation}
            $\tree$.init($\s_\start$) \;
            \While{\KwNot \method{STOPPING\_CONDITION}{} \label{alg_line:evs_sc}} {
                \tcc{node selection}
                $\s_\init \leftarrow$ $\tree$.\method{select\_node}{} \label{alg_line:evs_ns}
                
                \algspacing
                \tcc{micro-experience generation}
                $\s_\targ \leftarrow \varnothing$ \label{alg_line:evs_ss_start} \\
                \If{\method{ATTEMPT\_GOAL}{} = True} {
                    $\s_\targ \leftarrow \s_\goal$
                }
                $\langle\psi, \, \s_\targ\rangle \leftarrow$ \method{GENERATE\_SEGMENT}{$\s_\init$, $\s_\targ$, $\demo^\prime$} \label{alg_line:evs_ss_end}
                
                \algspacing
                \tcc{tree extension}
                \If{\method{EXTEND}{$\tree$, $\psi$, $\s_\init$, $\s_\targ$}  $\neq$ Failed \label{alg_line:evs_te_extend}} {
                    \If{\method{GOAL\_REACHED}{$\s_\targ$} \label{alg_line:evs_te_goal}} {
                        \KwReturn \method{PATH}{$\tree$} \label{alg_line:evs_te_return}
                    }    
                }
            }
        \end{algorithm}

        \aref{alg:evs} provides the pseudo-code of the uni-directional version of our planner. The algorithm seeks finding a continuous path from a start $\s_\start$ to a goal $\s_\goal$ configuration, given a related path experience $\demo$. The planner firstly maps the entire prior experience ${\nu: \demo \rightarrow \demo^\prime}$ onto the current planning problem (\lref{alg_line:evs_pro1}); note that the output of \method{GENERATE\_SEGMENT}{$\cdot$} (\aref{alg:segment}) is a segment $\psi$ that spans from ${\alpha_\ini=0}$ to ${\alpha_\fin=1}$, thus we rename it $\demo^\prime$. If $\demo^\prime$ is not valid (\lref{alg_line:evs_pro2}), the planner proceeds to exploit $\demo^\prime$ to generate task-relevant micro-experiences. The planner follows a three-step procedure (node selection, segment sampling, and tree extension) until the stopping condition is met (\lref{alg_line:evs_sc}). A node $\s_\init$ is selected from the tree \mbox{$\tree$ with probability ${P(node) = \frac{1}{w(node)+1}}$} (\lref{alg_line:evs_ns}), where $w(\cdot)$ is a weighting function that penalises the selection of a node according to the number of times that it has already been selected. This weighted sampling strategy seeks a uniform selection of all nodes over time, thus promoting a first depth exploration of the task phase~$\alpha$. 
        From the selected node $\s_\init$, the tree is expanded using segments that resemble those in the prior experience~$\demo^\prime$.
        With probability~$\tsg$, the expansion of the tree attempts to \textit{connect} $\s_\init$ with $\s_\goal$, whereas with probability $(1-\tsg)$ an \textit{explore} expansion is done towards a semi-random configuration $\s_\targ$ (\ltlref{alg_line:evs_ss_start}{alg_line:evs_ss_end}). \aref{alg:segment}, previously explained in \sref{sec:planner_segment}, details the extraction of suitable segments under these two different cases. The extracted segment is used to attempt expanding the tree (\lref{alg_line:evs_te_extend}) following \aref{alg:extend}. If the segment is valid,
        it is integrated into the tree.
        Note that, as discussed in \sref{sec:planner_segment}, the appended segment is a motion whose shape resembles that of the related micro-experiences in the prior experience, not a straight line. Finally, if the incorporated (valid) segment reaches the goal, the path is returned (\lalref{alg_line:evs_te_goal}{alg_line:evs_te_return}).

        \begin{algorithm}[t!]
            \SetInd{0.5em}{0.5em}
            \DontPrintSemicolon
            
            \caption{EXTEND($\tree$, $\psi$, $\s_\init$, $\s_\targ$)}
            \label{alg:extend}
            
            \nonl\textbf{Input:} \\
            \nonl\forceindent$\tree$: tree of previously generated micro-experiences \\
            \nonl\forceindent$\psi$: new generated micro-experience \\
            \nonl\forceindent$\s_\init$ and $\s_\targ$: start and end configuration-phase of $\psi$ \\
            
            \nonl\textbf{Output:} \\
            \nonl\forceindent outcome of the tree extension attempt \\
            
            
            \algspacing
            \If{\method{IS\_VALID}{$\psi$} \label{alg_line:extend_valid}} {
                $\tree$.\method{add\_vertex}{$\s_\targ$} \label{alg_line:extend_integrate_start} \;
                $\tree$.\method{add\_edge}{$\psi$, $\s_\init$, $\s_\targ$} \label{alg_line:extend_integrate_end} \;
                \KwReturn Advanced
            }
            \KwReturn Failed
        \end{algorithm}

    \subsection{Bi-directional ERT (ERTConnect) \label{sec:planner_bi}}
    
        The principles of leveraging from a prior experience by generation of task-relevant micro-experiences can also be employed in a bi-directional fashion. The proposed bi-directional planning scheme resembles, in spirit, that of the RRTConnect~\cite{kuffner2000rrt}, i.e., to simultaneously grow two trees, one from the start configuration and the other from the goal configuration, aiming to find a solution by connecting both trees. ERTConnect, however, includes the peculiarities of our experience-based planning approach. As shown in \aref{alg:evsconnect}, the planner firstly maps the prior path experience ${\nu: \demo \rightarrow \demo^\prime}$ onto the current planning problem (\lref{alg_line:evs_connect_pro1}). If $\demo^\prime$ is not valid (\lref{alg_line:evs_connect_pro2}), the planner proceeds to exploit $\demo^\prime$ to compute a solution. The planner simultaneously grows two trees, one rooted at $\s_\start$ and the other at $\s_\goal$ (\lalref{alg_line:evs_connect_setup1}{alg_line:evs_connect_setup2}). At each iteration, until the stopping criterion is met (\lref{alg_line:evsconnect_sc}), a node of the active tree $\tree_a$ is selected via weighted selection (\lref{alg_line:evsconnect_ns}) to \textit{explore} the space via a task-relevant micro-experience (\lref{alg_line:evsconnect_ss}). If the active tree is extended successfully with the generated segment (\lref{alg_line:evsconnect_extend1}), we first check whether the end of the motion has reached the other extreme (\lref{alg_line:evsconnect_return1_start}). This implies that a path has been found before the trees connected, either by $\tree_a$ reaching the root of $\tree_b$ or the other way around, in which case the path is returned as a solution (\lref{alg_line:evsconnect_return1_end}). Otherwise, the node of $\tree_b$ nearest to $\s_\targ$ is selected (\lref{alg_line:evsconnect_nearest}) to attempt to \textit{connect} both trees with a task-relevant segment (\lref{alg_line:evsconnect_ct}). For such nearest neighbour query, we consider the Euclidean distance between the configuration components (without the phase). If the extension is successful, the corresponding path is returned (\lalref{alg_line:evsconnect_extend2}{alg_line:evsconnect_return2}).
    
        \begin{algorithm}[t!]
            \SetInd{0.5em}{0.5em}
            \DontPrintSemicolon
            \caption{ERTConnect($\s_\start$, $\s_\goal$, $\demo$)} \label{alg:evsconnect}
            
            \nonl\textbf{Input:} \\
            \nonl\forceindent$\s_\start$ and $\s_\goal$: start and goal configuration-phase \\
            \nonl\forceindent$\demo$: prior experience \\
            
            \nonl\textbf{Output:} \\
            \nonl\forceindent$\xi$: collision-free path \\
            
            \algspacing
            \tcc{map $\demo$ onto current problem}
            $\langle\demo^\prime, \; \varnothing\rangle \leftarrow$ \method{GENERATE\_SEGMENT}{$\s_\start$, $\s_\goal$, $\demo$} \label{alg_line:evs_connect_pro1} \\
            \If{\method{IS\_VALID}{$\demo^\prime$} \label{alg_line:evs_connect_pro2}} {
                \KwReturn $\demo^\prime$
            }
            
            \algspacing
            
            \tcc{sampling-based $\demo^\prime$ exploitation}
            $\tree_a$.init($\s_\start$) \label{alg_line:evs_connect_setup1} \;
            $\tree_b$.init($\s_\goal$) \label{alg_line:evs_connect_setup2} \;
            \While{\KwNot \method{STOPPING\_CONDITION}{} \label{alg_line:evsconnect_sc}} {
                \tcc{node selection}
                $\s_\init \leftarrow$ $\tree_a$.\method{select\_node}{} \label{alg_line:evsconnect_ns} 
                
                \algspacing
                \tcc{micro-experience generation}
                $\langle\psi, \; \s_\targ\rangle \leftarrow$ \method{GENERATE\_SEGMENT}{$\s_\init$, $\varnothing$, $\demo^\prime$} \label{alg_line:evsconnect_ss}
                
                \algspacing
                
                \tcc{tree extension}
                \If{\method{EXTEND}{$\tree_a$, $\psi$, $\s_\init$, $\s_\targ$} $\neq$ Failed \label{alg_line:evsconnect_extend1}} {
                    \If{\method{OTHER\_EXTREME\_REACHED}{$\s_\targ$} \label{alg_line:evsconnect_return1_start}} {
                        \KwReturn \method{PATH}{$\tree_a$} \label{alg_line:evsconnect_return1_end}
                    }
                    
                    \algspacing
                    \tcc{micro-experience generation}
                    
                    $\s_\near \leftarrow$ $\tree_b$.\method{nearest\_neighbour}{$\s_\targ$} \label{alg_line:evsconnect_nearest}
                    

                    
                    \mbox{$\langle\psi, \; \s_\targ\rangle \leftarrow$ \method{GENERATE\_SEGMENT}{$\s_\near$, $\s_\targ$, $\demo^\prime$} \label{alg_line:evsconnect_ct}}

                    \algspacing
                    
                    \tcc{tree connection}
                    
                    \If{\method{EXTEND}{$\tree_b$, $\psi$, $\s_\near$, $\s_\targ$} $\neq$ Failed \label{alg_line:evsconnect_extend2}} {
                        \KwReturn \method{PATH}{$\tree_a$, $\tree_b$} \label{alg_line:evsconnect_return2}
                    }
                }
                \method{SWAP}{$\tree_a$, $\tree_b$}
            }
        \end{algorithm}

        
	\section{USING ERT and ERTConnect} \label{sec:using}
    In this section we discuss some details on using the proposed experience-driven random trees planners.
    
    \subsection{Selecting a Prior from a Library of Experiences \label{sec:preliminaries_selection}}
        A robot might have at its disposal a library $\mathcal{A}$ of task-relevant path experiences. 
        As our planners exploit a unique prior path experience $\demo$ to solve other instances of the same task, our current selection criteria ${\demo \in \mathcal{A}}$ is to pick the prior that resembles the current planning query the most.
        Intuitively, if a solution to the current planning problem lies in the neighbourhood of a prior experience, the invariant robotic constraints encoded in the experience itself, such as self-collisions and joint limits, are more likely to prevail and thus, ease the planner's computations. 
        Inspired by the experience selection in \cite{berenson2012robot}, we estimate such resemblance by ranking the experiences for their similarity to the start and goal of the planning query.
        This is, the prior experience $\demo$ selected to feed the proposed experience-based planner is such that:
        \begin{align}
            \demo = \arg \! \min_{{\demo}_i \in \mathcal{A}} \dist({\demo}_i(0), \; \q_\start) + \dist({\demo}_i(1), \; \q_\goal),
            \label{eq:path_selection}
        \end{align}
        where $\q_\start$ and $\q_\goal$ are the start and goal configurations of the current planning problem, and $\dist(\cdot)$ is a function that estimates the Euclidean distance between configurations in $\cspace$. 
        
        We verified the approach to select a unique prior from a library of experiences described in \eref{eq:path_selection} experimentally; despite it led to good results, the topic merits further attention.

    \subsection{Planner's Parameterisation \label{sec:planner_parameters}}
        Next, we review the parameters of the presented planners:
        \begin{itemize}
            \item $\tsg$ - probability of attempting to \textit{connect} the tree to the goal (only in ERT). This parameter should be set small to allow the planner \textit{explore} the space. Default: ${\tsg = 0.05}$.
            \item $\psl$ and $\psu$ - lower and upper phase span bounds of the extracted segments. Indirectly, these parameters delimit the length of the motions added in the tree to \textit{explore} the space. Default: ${\psl=0.05}$ and $\psu={0.1}$.
            \item $\tse$ - malleability bound to delimit the amount of morphing applied to the micro-experiences. Intuitively, this parameter defines a volume (tube) around $\demo^\prime$ where the planner can \textit{explore} for a solution.
            Default: ${\tse = \textbf{5}_{1\times n}}$ (large enough to cover the entire robot's kinematic range).
        \end{itemize}
        
        
        Our planners' default parameters are non-optimised for a particular planning problem, but left generic to succeed in many scenarios provided a relevant path experience. The planners' behaviour can be adjusted by tuning, namely, $\psl$, $\psu$ and $\tse$. As our planner discards motions that are not entirely valid, large phase spans might endanger the ability to build a tree. However, in scenarios with few obstacles, $\psl$ and $\psu$ can be set large to speed up planning computations. Also, lowering $\tse$ can speed up computations, as the tree growth would be more guided around the mapped experience~$\demo^\prime$. The lower $\tse$, the more dependant the planner is on the suitability of the provided experience as the probabilistic completeness is compromised. Knowing the level of dissimilarity between the experience and the current problem might aid in tuning $\tse$ to trade growth guidance and space exploration.
        
        
        

\section{EXPERIMENTAL EVALUATION} \label{sec:results}
    The proposed experience-guided random trees have been implemented in \acf{OMPL}~\cite{sucan2012the} and evaluated on the Fetch robot~\cite{wise2016fetch} in a shelf-stocking task. Fetch is a humanoid robot with a 7-\acs{DoF} arm attached on a sliding torso, thus requiring to plan in an 8-\acs{DoF} configuration space. Our experiments are designed to measure the generalisation capabilities of our planner in scenarios that involve different levels of dissimilarity between prior experiences and task instances (see \sref{sec:results_setup}). The considered task instances include synthetic and real-world scenarios (see \sref{sec:results_all}).

    \begin{figure*}[ht]
        \centering
        \begin{minipage}{.85\textwidth}
            \begin{minipage}{.31\textwidth}
                \centering
                \adjustbox{trim=-.1cm 0.3cm 0cm 0cm}{%
                  \includegraphics[height=0.8cm]{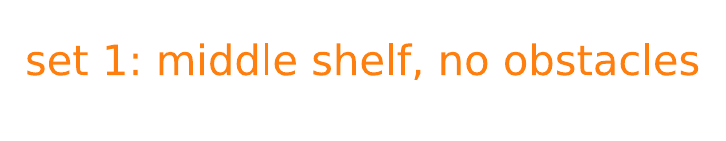}
                } \\
                \includegraphics[width=\textwidth]{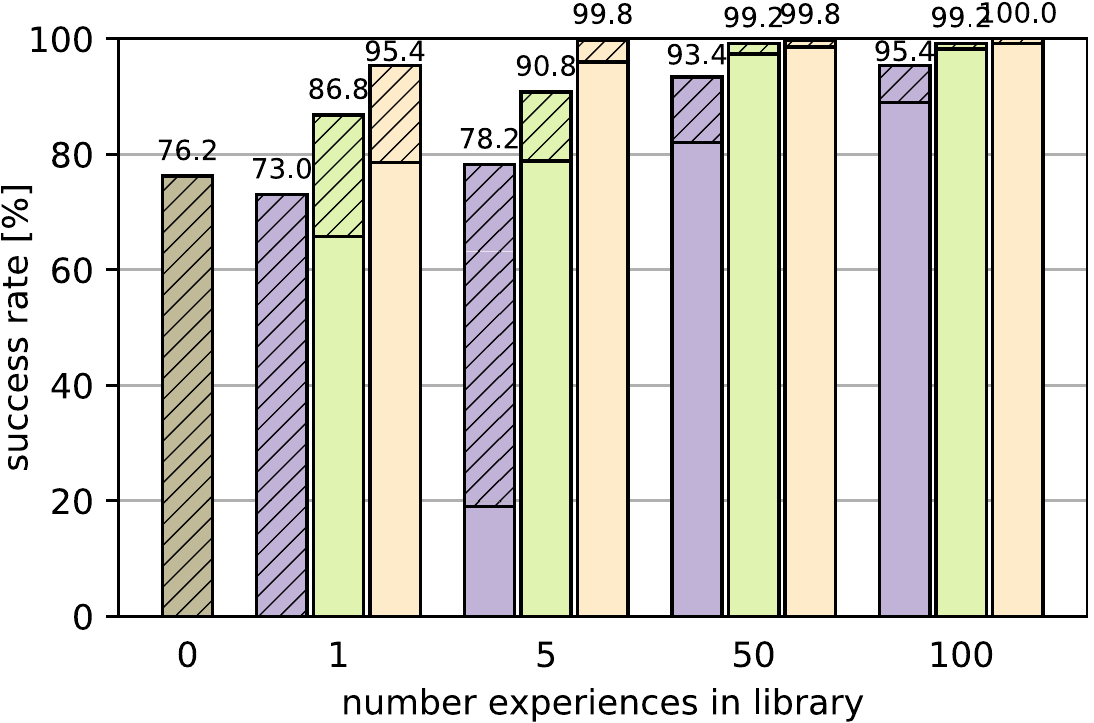} \\
                \includegraphics[width=\textwidth]{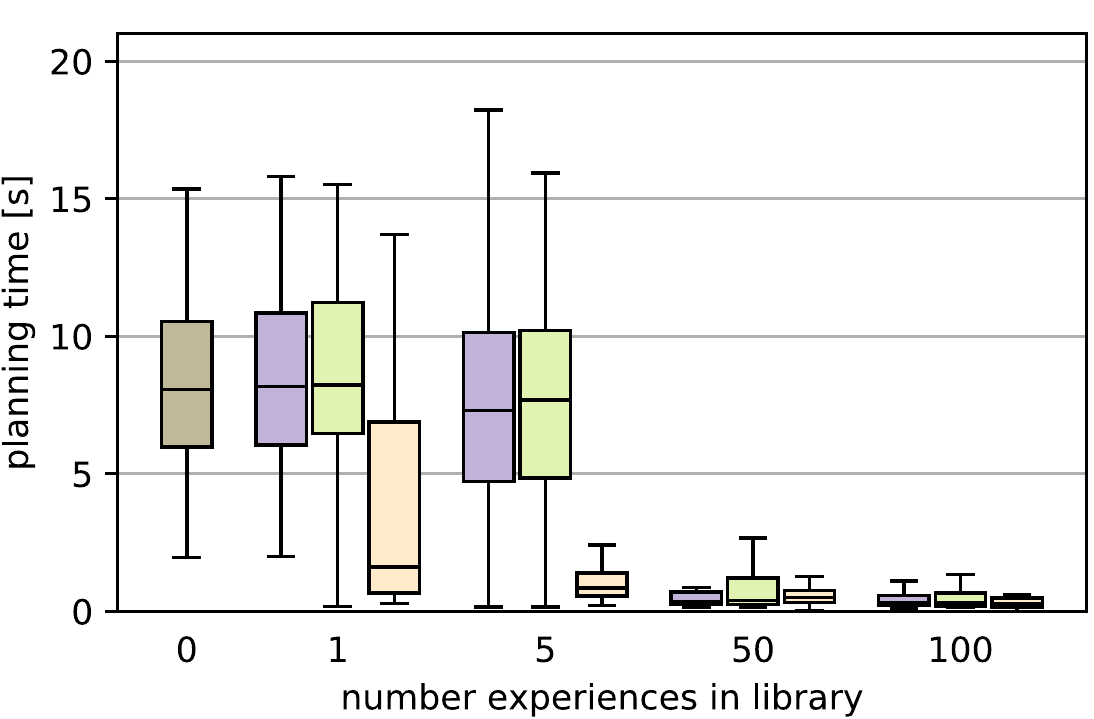}
            \end{minipage}%
            \;\!
            \begin{minipage}{.31\textwidth}
                \centering
                \adjustbox{trim=-.05cm 0.3cm 0cm 0cm}{%
                  \includegraphics[height=0.8cm]{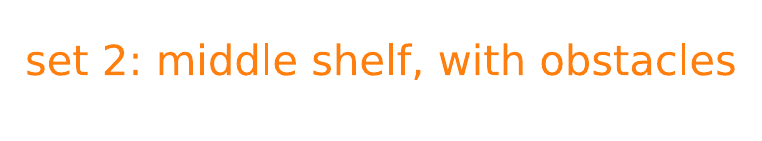}
                } \\
                \includegraphics[width=\textwidth]{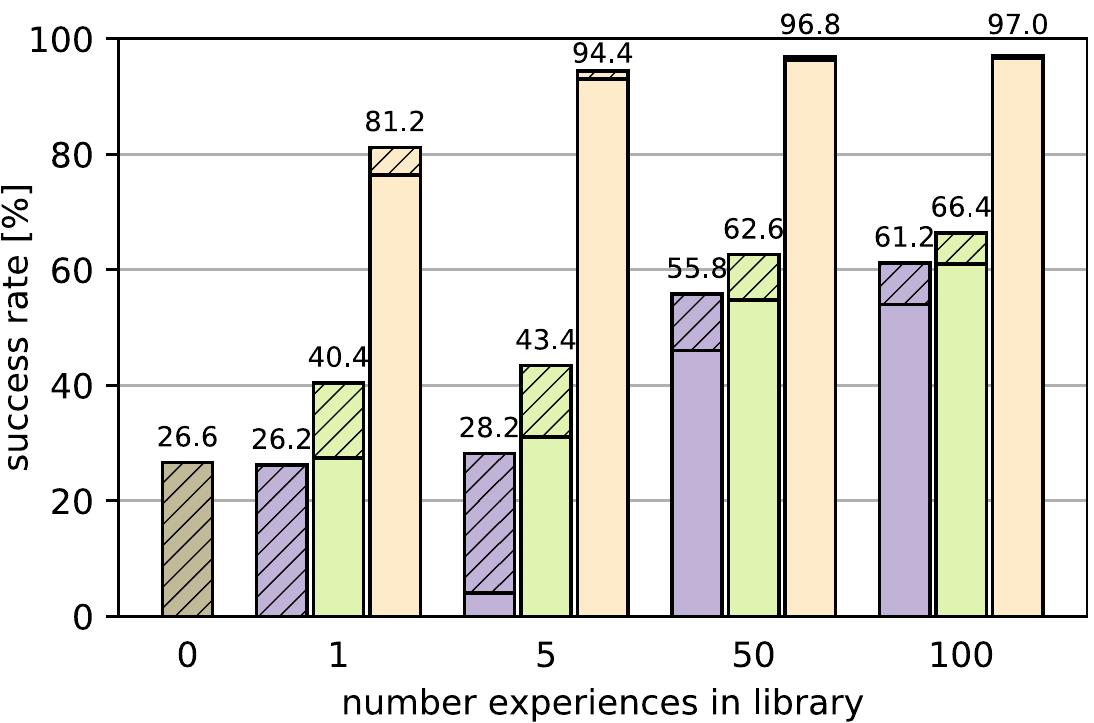} \\
                \includegraphics[width=\textwidth]{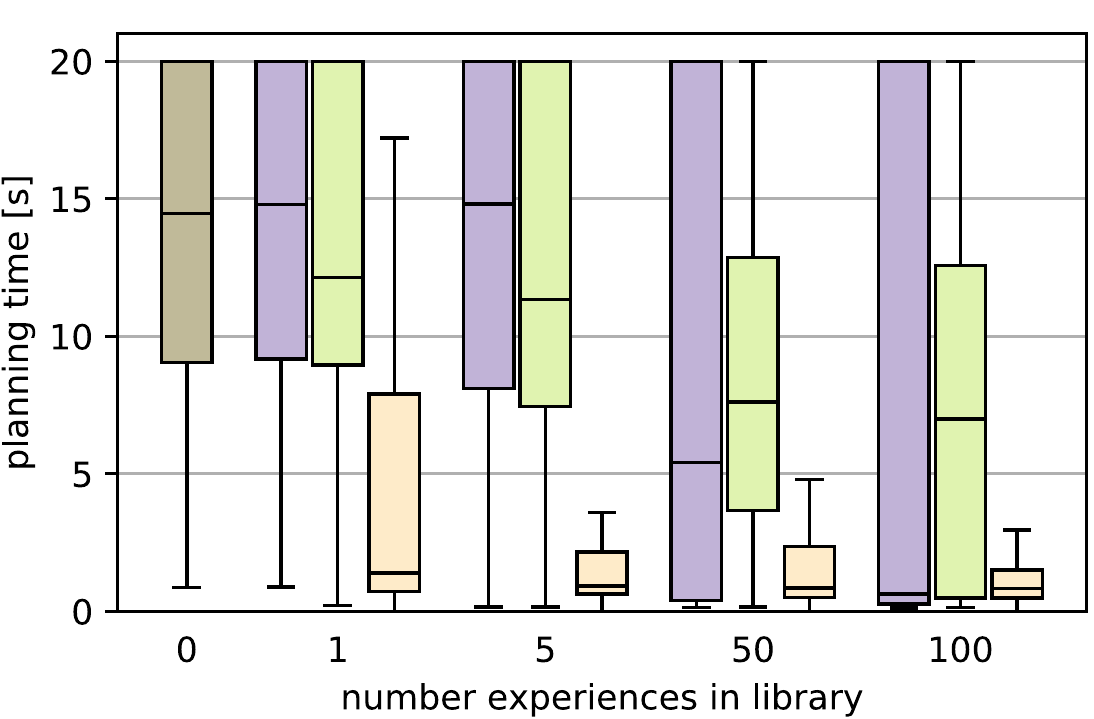}
            \end{minipage}%
            \;\!
            \begin{minipage}{.31\textwidth}
                \centering
                \adjustbox{trim=-.15cm 0.3cm 0cm 0cm}{%
                  \includegraphics[height=0.8cm]{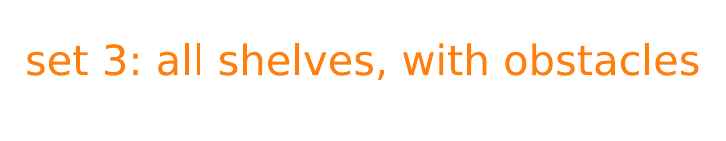}
                } \\
                \includegraphics[width=\textwidth]{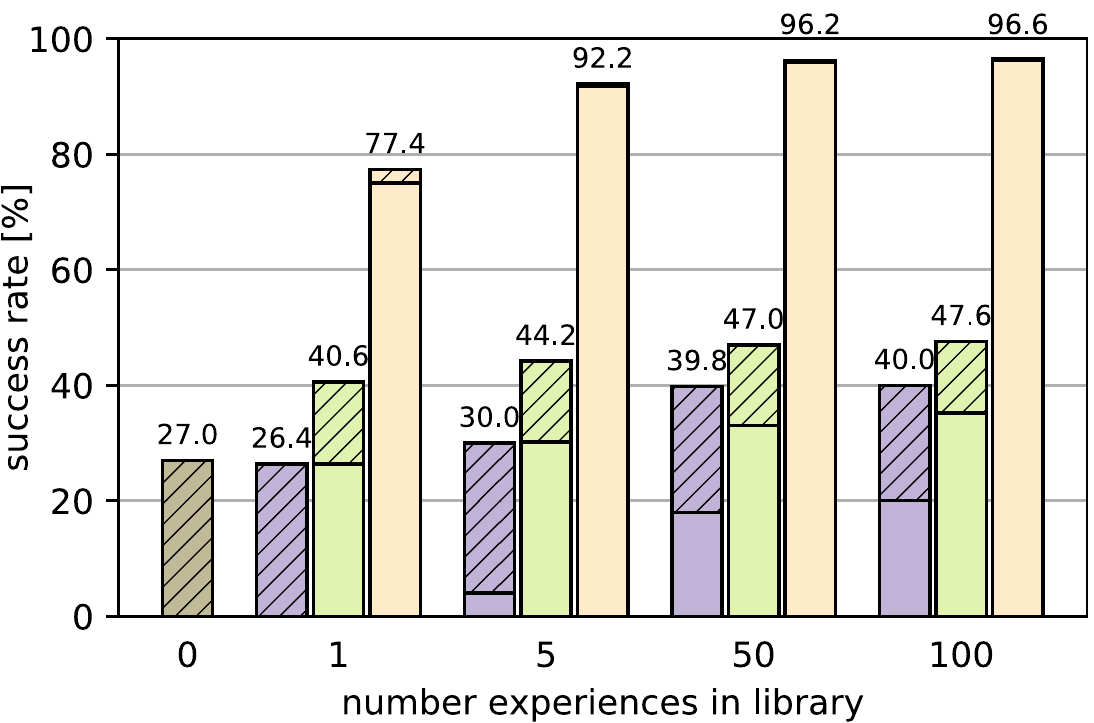} \\
                \includegraphics[width=\textwidth]{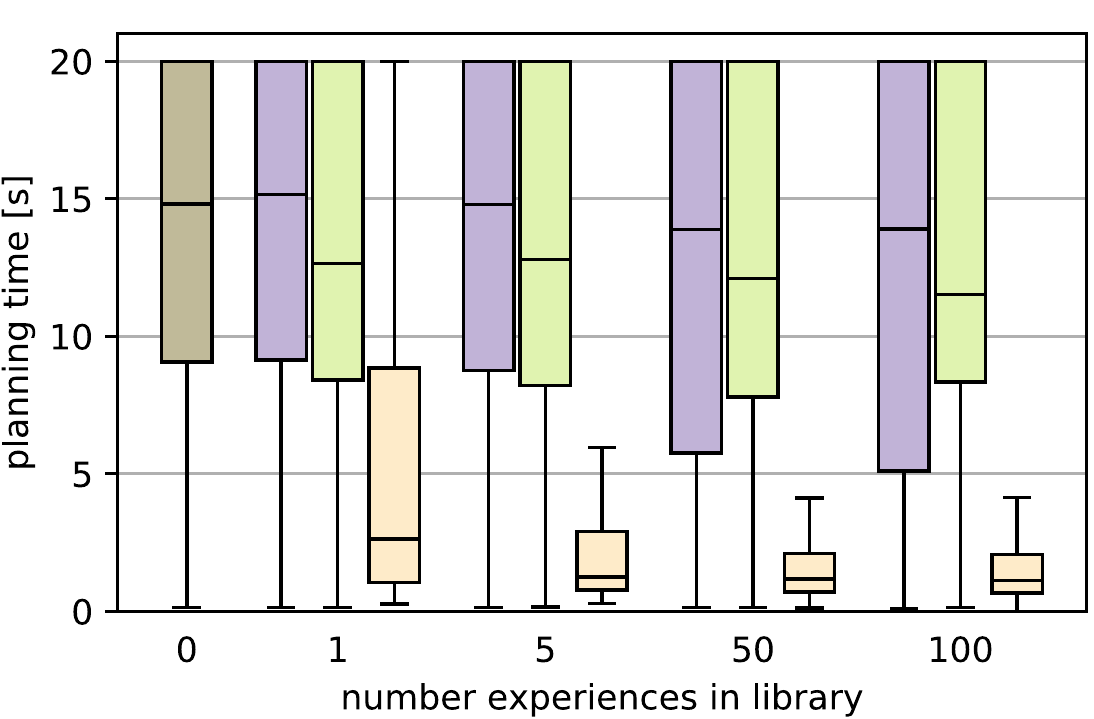}
            \end{minipage}%
        \end{minipage}%
        \hspace{-0.5cm}
        \begin{minipage}{.17\textwidth}
            \centering
            \vspace{0.2cm}
            \includegraphics[width=1\textwidth]{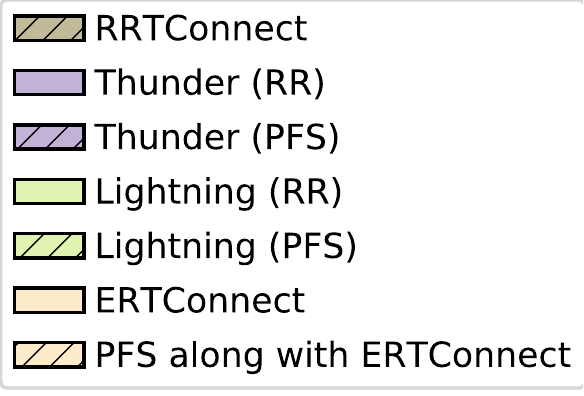}
            \includegraphics[width=\textwidth,trim={4cm 1cm 2cm 9cm},clip]{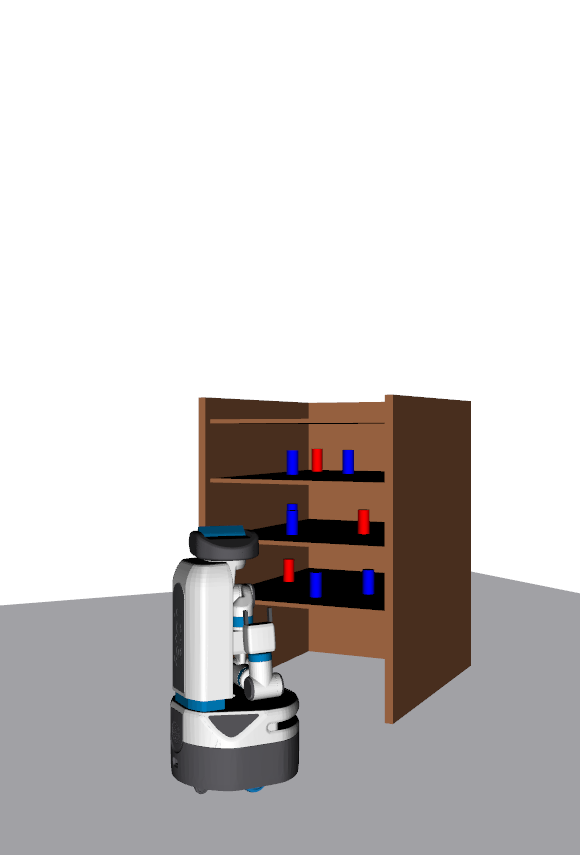} \\
        \end{minipage}%
        \caption{Success rate and solving time results for the benchmark on synthetic scenarios, where the Fetch Robot needs to reach a target object in the shelving unit subject to multiple variations of the task instances. 
        From left to middle-right column, case studies from less to more experience-instance dissimilarity: \textit{Set~1}, \textit{Set~2} and \textit{Set~3}.
        The picture on the right depicts a particular instance of \textit{Set~3} which, differently from the considered prior experiences, involves target objects (red cylinders) located at any of the three shelves, a different relative location of the shelving unit, as well as obstacles (blue cylinders).}
        \label{fig:test_synt_benchmark}
        \vspace{-0.35cm}
    \end{figure*}


    \subsection{Experimental Setup \label{sec:results_setup}}

        \modified{
        Our experimental setup considers a varied~instance~set~of the challenging problem of reaching a target object in a shelving unit, specifically in a synthetic 4-tier
        (see~\fref{fig:test_synt_benchmark}) and a narrower real 5-tier shelving unit (see~\fref{fig:test_real_benchmark}). Task instances in these scenarios not only present variability on the location of the shelving unit ($\pm90^\circ$ around the robot) and the robot's initial position ($\pm10\text{cm}$), but also on the location of the target object and the obstacles within the shelving unit.

        To further evaluate the generalisation capabilities of the proposed experience-based planner, we introduce some additional variability across the experimental setup. This is, we compute with RRTConnect a total of $100$ experiences from different task instances;
        them all at the synthetic shelving unit, with target objects in the middle shelf and no obstacles.
        These scenarios are discarded for the rest of the evaluation. Then, these experiences are used to evaluate the planner in four scenario sets that involve increasing dissimilarity levels between experiences and planning queries:
        \begin{itemize}
            \item \textit{Set~1}: $200$ instances with target objects in the middle shelf without obstacles (synthetic). Note that these instances resemble those used to compute experiences.
            \item \textit{Set~2}: $200$ instances with target objects in the middle shelf with the presence of obstacles (synthetic).
            \item \textit{Set~3}: $200$ instances with target objects in three different shelves with the presence of obstacles (synthetic).
            \item \textit{Set~4}: $120$ instances with target objects and obstacles in the middle shelf (real-world).
        \end{itemize}


        The four sets of task instances are used to benchmark our bi-directional ERTConnect planner against RRTConnect~\cite{kuffner2000rrt} and the most representative experience-based planners that employ motions as prior information of the task, i.e., Thunder~\cite{coleman2015experience} and Lightning~\cite{berenson2012robot}.
        Note that these two frameworks are double-threaded with a bi-directional `retrieve and repair' (RR) and `plan from scratch' (PFS) module. Similarly, for a fair comparison, we embed our ERTConnect in a double-threaded framework which runs RRTConnect in parallel to PFS. The reported results indicate the contribution of the RR (plain bar) and PFS (stripped bar) modules in solving the planning queries separately, and the required planning time jointly (plain bar).

        In this work's context, where we consider novel tasks instances in varied scenarios, optimising each planner's parameters across queries is not possible. Optimal parametrisation requires extensive testing in each scenario set, and thus knowing the scenarios in advance, among other planning aspects.
        Therefore, all planners are used in their default OMPL settings, and ours is set to the non-optimised default parameters specified in \sref{sec:planner_parameters}. The benchmark is run on an Intel i7 Linux machine with 4 3.6GHz cores and 16GB of RAM. The performance of the three experience-based planners in each instance set is assessed under libraries with ${\{1, 5, 50, 100\}}$ prior experiences. The experiences provided to each planner are the same. Each query is repeated $50$ times with a planning timeout of $20$ seconds. All in all, the conducted benchmark involves a total of $468{,}000$ planning queries.

    \subsection{Results on Synthetic and Real-world Scenarios \label{sec:results_all}}

        The results of the benchmark on \textit{Set~1}, \textit{Set~2} and \textit{Set~3} are summarised in \fref{fig:test_synt_benchmark}, whilst those in the real-world \textit{Set~4} are depicted in \fref{fig:test_real_benchmark}. As it can be observed, provided a high number of experiences that are close to the current planning problem (i.e., \textit{Set~1} with the library of $100$ experiences), all planners achieve a high success rate with solving time of the order of milliseconds.
        This behaviour is expected as, given the experience-query similarity and the library size, it is likely that there exists a prior experience that nearly resembles the \mbox{current query, thus involving minimum repairing.}

        \begin{figure}[b!]
            \centering
            \begin{minipage}{0.55\columnwidth}
                \centering
                \adjustbox{trim=-.3cm 0.35cm 0cm 0cm}{%
                 \includegraphics[height=0.8cm]{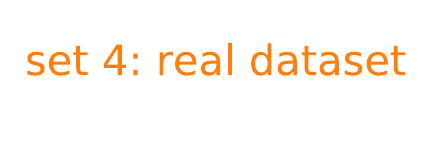}
                } \\
                \includegraphics[width=\columnwidth]{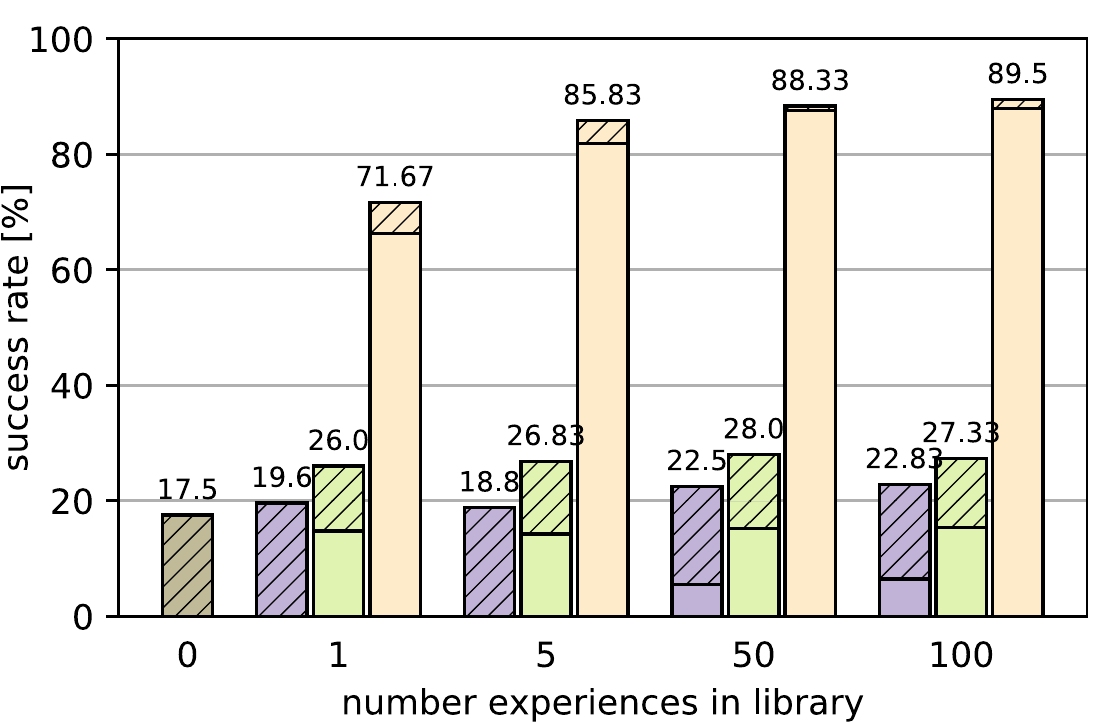} \\
                \includegraphics[width=\columnwidth]{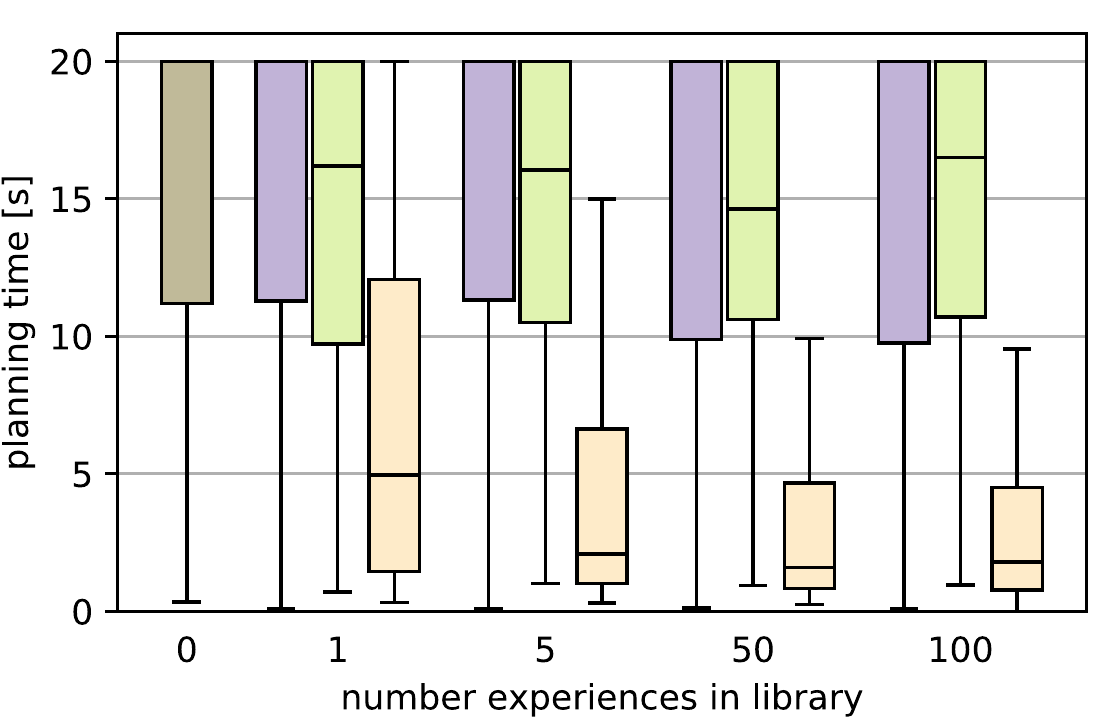}
            \end{minipage}%
            \hspace{0.25cm}
            \begin{minipage}{.35\columnwidth}
                \centering
                \vspace{0.15cm}
                \includegraphics[width=1\columnwidth]{./Figures/test_benchmark_legend_xxx.pdf} \\
                \vspace{0.1cm}
                \includegraphics[width=0.94\columnwidth]{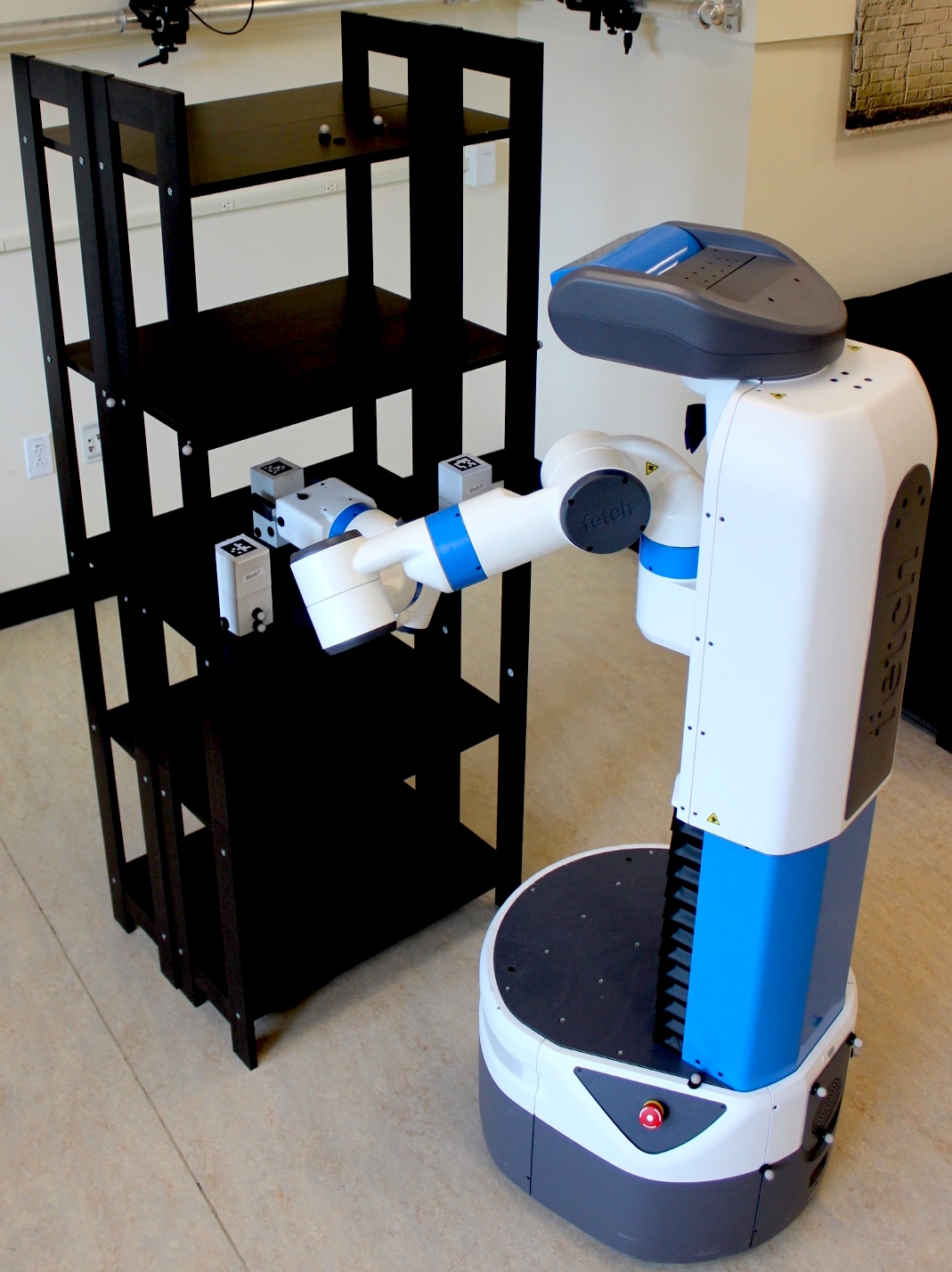}
            \end{minipage}%
            \caption{Success rate and solving time results for the benchmark on real-world scenarios (\textit{Set~4}), where the Fetch Robot needs to reach a target object by generalising prior experiences to a narrower shelving unit geometry, to different locations of the shelving unit, robot's initial position \mbox{and target object, as well as to the presence of obstacles.}}
            \label{fig:test_real_benchmark}
        \end{figure}

        As the dissimilarity between experiences and queries increases, experience-based planners need to generalise the prior information more broadly to succeed. Intuitively, the need of generalisation arises when a reduced number of demonstrations in the library needs to cover varied task instances (\mbox{x-axis} within each experimental set), or when the current planning requirements differ significantly from the set of available experiences (variability across experimental sets). The outcome of our benchmark points out that the performance of Thunder's
        RR module drops abruptly by either dissimilarity factor, whereas the Lighting's RR is not as affected by the lack of experiences as it is when dealing with significantly different task instances. The poor generalisation of these frameworks across instances is
        due to the rigid usage of prior experiences. Our approach, instead, by leveraging experiences in a malleable way, achieves a success rate and solving time that significantly improves that of Lightning and Thunder.

        The importance of generalising prior information is particularly noticeable in the real-world task instances in \textit{Set~4}, where the queries differ from the experiences not only on the location of the shelving unit and the target object, but also on the narrower geometry of the whole shelving unit, the height of the shelf where the target object is located at, and the presence of obstacles. Under these challenging task variations and when accounting with only one demonstration, our approach outperforms by a factor of approximately $3.7$ and $37.1$ times the
        RR module of the Lightning and Thunder frameworks, respectively. Similarly, those frameworks are respectively outperformed by our approach by a factor of $4.7$ and $8.9$ times when considering $100$ experiences. 
        Notably, while these planners time out most of the trials, our method required only a quarter (with one demonstration) and less than an eight (with multiple demonstrations) \mbox{of the planning time budget to find a solution.}


        The ability to generalise prior information not only limits the level of experience-query dissimilarity that a planner can cope with, but also the number of experiences that are required to achieve high performance. As an example, providing $50$ experiences to Thunder's RR, $5$ to Lightning's  RR and $1$ to our ERTConnect leads to approximately the same success rate of $80\%$ in \textit{Set~1}. Achieving such a performance in the other experimental sets with Lightning and Thunder is not possible even with a library of $100$ experiences, whereas our approach surpasses such performance when selecting a unique experience from a library of only $5$ experiences. This implies that our ERTConnect planner, by generalising prior experiences more efficiently, significantly outperforms current experienced-based planners using libraries of motions in the literature even when provided with notably fewer experiences.
        }

    \section{DISCUSSION} \label{sec:conclusions}
    
    In this manuscript, we have presented two new experience-based planners: the uni-directional experience-driven random tress (ERT) and the bi-directional ERT (ERTConenct). These two methods are tree sampling-based planners that iteratively exploit a single prior path experience to ease the capture of connectivity of the space. At each iteration, a segment of the prior is extracted and semi-randomly morphed to generate a task-relevant motion. The obtained motions are sequentially concatenated to compose a task-relevant tree, such that a trace along the edges constitutes a solution to a given task-related planning problem.
    Thorough experimentation against current experienced-based planners using libraries of motions in the literature~\cite{berenson2012robot,coleman2015experience} demonstrates our planner's significant superior performance in a wide range of task instances.
    
    
    \modified{We have shown that, similarly to related work~\cite{berenson2012robot,coleman2015experience}, our planner can be used in parallel with a planning from scratch strategy to guarantee probabilistic completeness.} 
    Therefore, when multi-threading is an option, a planning from scratch thread should be considered, as well as multiple instantiations of our planner with a set of varied experiences that maximises space coverage.
    In the future, we plan to explore the convenience of different transformation supports to infer relevant micro-experiences subject to intrinsic task and robot constraints; for instance, early tests demonstrate our planners' suitability to leverage experiences in $\mathrm{SO}(3)$ using quaternions. Another promising line for future work is the extension of our planner to leverage multiple experiences simultaneously, such that the local exploration is conducted with the most suitable segment in the library. Likewise, such a strategy would potentially allow the planner adapting to changes in the planning context, e.g., dynamic obstacles and moving goal configurations, as well as solving novel tasks by combining experiences of multiple different tasks.
    
    



    
    
    



    \section*{ACKNOWLEDGMENTS}
    The authors thank Zachary Kingston for~his~support~in~integrating our planner into MoveIt, and Carlos Quintero 
    for his help on the final experiments. Also, the authors are grateful to Paola Ard{\'o}n for \mbox{valuable discussions and suggestions}.

	\bibliographystyle{ieeetr}
    \bibliography{refs}

\begin{thebibliography}{10}

\bibitem{meier2018online}
F.~Meier, D.~Kappler, and S.~Schaal, ``Online learning of a memory for learning
  rates,'' in {\em 2018 IEEE International Conference on Robotics and
  Automation}, pp.~2425--2432, IEEE, 2018.

\bibitem{pairet2019blearning}
{\`E}.~Pairet, P.~Ard{\'o}n, M.~Mistry, and Y.~Petillot, ``Learning
  generalizable coupling terms for obstacle avoidance via low-dimensional
  geometric descriptors,'' {\em IEEE Robotics and Automation Letters}, vol.~4,
  no.~4, pp.~3979--3986, 2019.

\bibitem{stark2019experience}
S.~Stark, J.~Peters, and E.~Rueckert, ``Experience reuse with probabilistic
  movement primitives,'' in {\em IEEE/RSJ International Conference on
  Intelligent Robots and Systems}, pp.~1210--1217, IEEE, 2019.

\bibitem{pairet2019alearning}
{\`E}.~Pairet, P.~Ard{\'o}n, M.~Mistry, and Y.~Petillot, ``Learning and
  composing primitive skills for dual-arm manipulation,'' in {\em Annual
  Conference Towards Autonomous Robotic Systems}, pp.~65--77, Springer, 2019.

\bibitem{ravichandar2020recent}
H.~Ravichandar, A.~S. Polydoros, S.~Chernova, and A.~Billard, ``Recent advances
  in robot learning from demonstration,'' {\em Annual Review of Control,
  Robotics, and Autonomous Systems}, vol.~3, 2020.

\bibitem{zucker2008adaptive}
M.~Zucker, J.~Kuffner, and J.~A. Bagnell, ``Adaptive workspace biasing for
  sampling-based planners,'' in {\em 2008 IEEE International Conference on
  Robotics and Automation}, pp.~3757--3762, IEEE, 2008.

\bibitem{ichter2018learning}
B.~Ichter, J.~Harrison, and M.~Pavone, ``Learning sampling distributions for
  robot motion planning,'' in {\em 2018 IEEE International Conference on
  Robotics and Automation}, pp.~7087--7094, IEEE, 2018.

\bibitem{lehner2018repetition}
P.~Lehner and A.~Albu-Sch{\"a}ffer, ``The repetition roadmap for repetitive
  constrained motion planning,'' {\em IEEE Robotics and Automation Letters},
  vol.~3, no.~4, pp.~3884--3891, 2018.

\bibitem{chamzas2019using}
C.~Chamzas, A.~Shrivastava, and L.~E. Kavraki, ``Using local experiences for
  global motion planning,'' in {\em 2019 International Conference on Robotics
  and Automation}, pp.~8606--8612, IEEE, 2019.

\bibitem{molina2020learn}
D.~Molina, K.~Kumar, and S.~Srivastava, ``Learn and link: learning critical
  regions for efficient planning,'' in {\em {IEEE} International Conference on
  Robotics and Automation}, 2020.

\bibitem{berenson2012robot}
D.~Berenson, P.~Abbeel, and K.~Goldberg, ``A robot path planning framework that
  learns from experience,'' in {\em 2012 IEEE International Conference on
  Robotics and Automation}, pp.~3671--3678, IEEE, 2012.

\bibitem{jetchev2013fast}
N.~Jetchev and M.~Toussaint, ``Fast motion planning from experience: trajectory
  prediction for speeding up movement generation,'' {\em Autonomous Robots},
  vol.~34, no.~1-2, pp.~111--127, 2013.

\bibitem{phillips2012graphs}
M.~Phillips, B.~J. Cohen, S.~Chitta, and M.~Likhachev, ``E-graphs:
  bootstrapping planning with experience graphs.,'' in {\em Robotics: Science
  and Systems}, 2012.

\bibitem{coleman2015experience}
D.~Coleman, I.~A. {\c{S}}ucan, M.~Moll, K.~Okada, and N.~Correll,
  ``Experience-based planning with sparse roadmap spanners,'' in {\em 2015 IEEE
  International Conference on Robotics and Automation}, pp.~900--905, IEEE,
  2015.

\bibitem{wang2019motion}
Y.~Wang, K.~Harada, and W.~Wan, ``Motion planning through demonstration to deal
  with complex motions in assembly process,'' in {\em 2019 IEEE-RAS 19th
  International Conference on Humanoid Robots}, pp.~544--550, IEEE, 2019.

\bibitem{delpreto2020helping}
J.~DelPreto, J.~I. Lipton, L.~Sanneman, A.~J. Fay, C.~Fourie, C.~Choi, and
  D.~Rus, ``Helping robots learn: a human-robot master-apprentice model using
  demonstrations via virtual reality teleoperation,'' in {\em {IEEE}
  International Conference on Robotics and Automation}, 2020.

\bibitem{ardon2020self}
P.~Ard{\'o}n, {\`E}.~Pairet, Y.~Petillot, R.~P. Petrick, S.~Ramamoorthy, and
  K.~S. Lohan, ``Self-assessment of grasp affordance transfer,'' in {\em
  IEEE/RSJ International Conference on Intelligent Robots and Systems}, IEEE,
  2020.

\bibitem{hsu1997path}
D.~Hsu, J.-C. Latombe, and R.~Motwani, ``Path planning in expansive
  configuration spaces,'' in {\em Proceedings of International Conference on
  Robotics and Automation}, vol.~3, pp.~2719--2726, IEEE, 1997.

\bibitem{kuffner2000rrt}
J.~J. Kuffner and S.~M. LaValle, ``{RRT-connect: an efficient approach to
  single-query path planning},'' in {\em 2000 International Conference on
  Robotics and Automation}, vol.~2, pp.~995--1001, IEEE, 2000.

\bibitem{sucan2012the}
I.~A. {\c{S}}ucan, M.~Moll, and L.~E. Kavraki, ``{The {O}pen {M}otion
  {P}lanning {L}ibrary},'' {\em {IEEE} Robotics \& Automation Magazine},
  vol.~19, pp.~72--82, December 2012.

\bibitem{wise2016fetch}
M.~Wise, M.~Ferguson, D.~King, E.~Diehr, and D.~Dymesich, ``{Fetch and Freight:
  standard platforms for service robot applications},'' in {\em Workshop on
  autonomous mobile service robots}, 2016.

\end{thebibliography}
\end{document}